%% file: main.tex
\documentclass{article}

\PassOptionsToPackage{numbers,sort&compress,round}{natbib}

\usepackage[preprint]{neurips_2025}   
\usepackage{times}                    
\usepackage[T1]{fontenc}
\usepackage[utf8]{inputenc}
\usepackage{amsmath,amssymb,booktabs,tabularx,graphicx,microtype,longtable}
\usepackage{xcolor,float}

\definecolor{RevisionBlue}{gray}{0}
\newcommand{\rev}[1]{#1}

\usepackage{hyperref,url}
\IfFileExists{xurl.sty}{\usepackage{xurl}}{}
\hypersetup{
  hidelinks,
  pdftitle={Layer--Readout Interfaces in EEG Foundation Models: Predictive Accessibility and Cross-Dataset Transfer},
  pdfauthor={Yanlin Fu, Wei Liu, Liming Zhao},
  pdfkeywords={EEG foundation models, probing, layer selection, readout,
               cross-dataset transfer}
}
\title{Layer--Readout Interfaces in EEG Foundation Models:\\
       Predictive Accessibility and Cross-Dataset Transfer}
\author{%
  Yanlin Fu \qquad Wei Liu \qquad Liming Zhao \\[0.4em]
  QMindTech, Shanghai, China
}

\newcommand{\udepth}{\Delta_{\rm depth}}
\newcommand{\uinterface}{\Delta_{\rm joint}}

\begin{document}
\maketitle

\input{sections/00_abstract}
\input{sections/01_introduction}
\input{sections/02_related_work}
\input{sections/03_methods}
\input{sections/04_results}
\input{sections/05_discussion}
\input{sections/06_conclusion}

\bibliographystyle{unsrtnat}          
\bibliography{references}

%
\clearpage
\appendix
\input{sections/supplementary}

\clearpage      
\end{document}

%% file: sections/00_abstract.tex
\begin{abstract}
Intermediate layers often outperform final layers when foundation models are used as frozen feature extractors, but the source of this advantage is unclear. We study this question in EEG foundation models by separating three factors: how representations are read out, whether intermediate layers \rev{provide predictive utility beyond the final-layer interface}, and whether useful depth structure transfers across datasets. Across five models and three tasks, improving the final-layer readout consistently recovers substantial predictive performance, yet intermediate layers remain advantageous \rev{within frozen extraction}. We also find that layers selected on one dataset remain useful on another \rev{across five dataset pairs and four task families}, and that a compact set of source-derived task directions retains predictive value after transfer \rev{in sleep staging}. Together, these results show that \rev{predictive utility varies across depth, and its usefulness depends on how representations are read out and transferred.
More crucially, source-selected layers provide a reusable feature interface across datasets, reducing the need for target-side layer search while outperforming final-layer baselines on
target datasets.}
\end{abstract}

%% file: sections/01_introduction.tex
\section{Introduction}
Electroencephalography foundation models (EEG-FMs) have recently emerged as a powerful paradigm for neural decoding, leveraging large-scale pre-training to learn rich, transferable brain representations across diverse paradigms and recording configurations. However, when evaluating these frozen encoders for downstream tasks, probes on intermediate layers frequently outperform those on the final layer. \rev{This observation raises a fundamental question: how much of the apparent final-layer disadvantage
reflects differences in predictive representations, and how much arises from how hidden tokens are aggregated?} In the context of EEG FMs, the answer is tightly coupled with the readout mechanism. Standard global averaging over spatial channels and temporal steps deliberately discards structured dynamics, whereas channel- or dispersion-preserving summaries retain them. Consequently, a weak final-layer baseline may be less a limitation of the representation itself than a symptom of a suboptimal readout.

While recent work highlights that layer selection and pooling strategies profoundly impact frozen representation decoding across vision and biosignal domains \citep{evci2022head2toe,sirca2026beyond}, downstream evaluations remain acutely sensitive to choices in probing protocols and classification heads \citep{kontras2026neuroatlas,wang2026audit,lahiri2026prism}. \rev{\cite{evci2022head2toe} selects intermediate features using target-task data, whereas \cite{sirca2026beyond} examines pooling and block-wise probing in EEG-FMs. Neither directly tests whether a layer-readout interface selected on one downstream EEG dataset remains useful on another without repeating target-side layer selection.} These insights leave two critical questions unresolved: Do intermediate layers still offer \rev{additional predictive utility}
once the final-layer readout is selected?
And can \rev{a source-selected layer-readout interface be reused across other datasets}?

\begin{figure}[!htp]
\centering
\includegraphics[width=\linewidth]{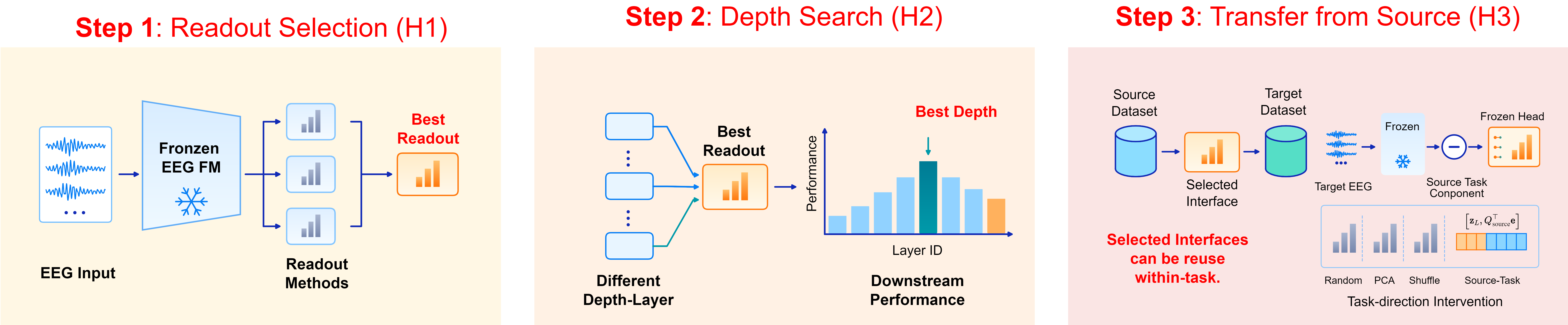}
\caption{\rev{\textbf{Evaluation framework for frozen EEG
foundation models.}
\textbf{Step 1 (H1):} Select the final-layer readout.
\textbf{Step 2 (H2):} Select depth under the fixed
readout.
\textbf{Step 3 (H3):} Reuse source-selected interfaces
across datasets and evaluate source-derived task
directions through intervention.}}
\label{fig:story}
\end{figure}

We structure our investigation around three hypotheses, \rev{and our findings to the hypotheses are} as illustrated in Figure~\ref{fig:story}. First, we hypothesize that readout mismatch accounts for a substantial portion of the final layer's apparent performance gap (\textbf{H1}). Second, intermediate layers \rev{may provide additional
predictive utility after fixing the final-layer readout} (\textbf{H2}). Third, \rev{we test whether a source-selected layer-readout
interface remains useful on another dataset of the same task, and whether source-derived task directions retain predictive utility after transfer} (\textbf{H3}). We systematically test these hypotheses by isolating
readout selection from depth selection,
\rev{assessing dimensionality-dependent residual utility},
and validating cross-dataset transfer
\rev{through explicit layer controls and subspace interventions}.

By evaluating five representative foundation models across motor imagery and two event-related potential (ERP) paradigms, we first reproduce the widespread empirical finding that intermediate layers often outperform final-layer representations. Selecting the readout improves final-layer prediction in every condition, and selecting the layer yields further gains in most conditions. \rev{These interface choices also alter the relative
ranking of foundation models. We further use cross-fitted residual augmentation
to assess incremental predictive utility and its
sensitivity to feature dimensionality.}

Finally, source-selected layers remain useful in
within-task cross-dataset transfer for sleep staging
and motor imagery, as well as SSVEP and P300,
\rev{outperforming final-layer baselines in all ten
CBraMod/REVE comparisons across five dataset pairs}. Source residual directions also improve sleep transfer across three model families. In sleep staging, removing these source task directions drops target classifier performance significantly more than matched controls, though refitting the classification head recovers most of the accuracy. Together, these results separate readout artifacts from \rev{depth-dependent predictive utility and support selecting the final-layer readout before evaluating depth, then reusing the source-selected interface across datasets.}

%% file: sections/02_related_work.tex
\section{Related Work}
\subsection{EEG foundation models and their \rev{evaluations}}
Early work on EEG pretraining laid the foundation with two main approaches: BENDR introduced contrastive self-supervised learning \citep{kostas2021bendr}, while MAEEG focused on reconstructing masked EEG features \citep{chien2022maeeg}. Later models expanded on these ideas. BIOT introduced support for varying channel setups \citep{yang2023biot}, and both EEGPT and EEG2Rep trained on masked signals by using learning objectives directly in the representation space \citep{wang2024eegpt,foumani2024eeg2rep}.

The five models we examine \rev{vary in architecture,
tokenization, and pretraining objectives}. LaBraM uses spectrum prediction to learn discrete codes across different datasets \citep{jiang2024labram}. CBraMod separates spatial and temporal information using a cross-attention mechanism paired with masked patch reconstruction \citep{wang2025cbramod}. REVE combines masked autoencoding with positional embeddings to handle flexible electrode setups and varying signal lengths \citep{elouahidi2025reve}. LUNA uses learned queries to compress channels into a fixed set of hidden tokens, decoupling feature representations from the physical layout of the electrodes \citep{doner2025luna}. Finally, CSBrain divides signals across temporal windows and brain regions using structured sparse attention \citep{zhou2025csbrain}. Because these designs preserve different aspects of the input signal, both the choice of intermediate layer and how its outputs are summarized directly affect what information reaches the downstream classifier.

Standard evaluations usually test these models either by training a simple linear classifier on top of frozen features or by fine-tuning the whole network on a new task. Recent benchmarks have standardized probing on frozen encoders \citep{kontras2026neuroatlas} and compared how models handle different adaptation strategies and signal perturbations \citep{wang2026audit}. Classifier architecture choice also introduces variability, though most prior comparisons focus on full fine-tuning scenarios \citep{lahiri2026prism}. \rev{In our primary analysis,} we keep the encoder fixed and focus on how feature extraction varies across both layer depth and readout design, benchmarked against the standard final-layer baseline.

\subsection{Extracting information from intermediate layers}
It is well established that feature usefulness varies with model depth and depends heavily on how closely the pretraining task aligns with the downstream target \citep{yosinski2014transferable}. Training independent linear classifiers at each layer offers a straightforward way to trace how class separability evolves across a model \citep{alain2016probes}. For EEG models specifically, preserving the underlying token layout often improves downstream accuracy: simply flattening tokens frequently works better than averaging them, and intermediate blocks can contain rich, useful representations \citep{sirca2026beyond}. Combining features from multiple layers can similarly boost performance on new tasks \citep{evci2022head2toe}, a trend also observed in time-series and audio models \citep{feofanov2026mantis,rauch2026unmute}. These insights highlight why we need to evaluate not just which layer to extract features from, but also how those features are summarized. Building on this, we \rev{examine how readout and depth jointly affect predictive utility, and whether source-selected layer-readout interfaces remain useful across datasets}.

Separating a practical feature summary from a genuinely rich representation is also a key challenge in language modeling. For instance, pooling attention across layers can improve sentence embeddings \citep{oh2022lastlayer}. Conversely, certain dominant feature dimensions can artificially skew similarity metrics without reflecting actual model behavior \citep{timkey2021rogue}, often driven by frequency artifacts in the inputs \citep{puccetti2022outlier, fuster2022anisotropy}. These observations demonstrate that conclusions about representation geometry or feature quality are highly sensitive to the choice of readout.

\subsection{Evaluating feature utility and representations}
A classifier's accuracy reflects not only the quality of the underlying representation, but also the classifier's own capacity to learn. Prior studies use synthetic control tasks or measure the learning effort required to extract information to avoid overestimating representation quality \citep{hewitt2019controls,voita2020mdl}. Merely being able to decode a feature does not mean the model actually relies on it to make predictions \citep{belinkov2022probing}.
To test whether features are functionally important, we can selectively remove specific information from the representation and measure the impact on downstream task performance \citep{elazar2021amnesic, ravfogel2020inlp, belrose2023leace}. In EEG research, feature removal has typically focused on explicit, predefined physiological properties \citep{tang2026capture}. Here, we adopt a different approach: we isolate \rev{residual directions under a specified final-to-intermediate mapping and feature dimension, and test their predictive utility across datasets using matched augmentation and deletion controls}.

%% file: sections/03_methods.tex
\section{Methods}

\subsection{Representations and Readout Families}
Let $H_\ell(x)$ denote the hidden representations at layer $\ell \in \{1, \dots, L\}$ of a frozen foundation model encoder given an EEG input $x$, where $L$ represents the final layer. Downstream classifiers access these representations through a readout function $r(\cdot)$, yielding a fixed-dimensional feature vector $z_{\ell,r}(x) = r(H_\ell(x))$.

To evaluate \rev{how feature aggregation affects predictive utility}, we define a primary family of fixed readouts $\mathcal{R}$ for channel-time tensors. This includes global mean pooling, global mean plus temporal standard deviation, channel-wise mean pooling, channel-wise mean with temporal standard deviation, and temporal binning. For architectures lacking a physical channel dimension (e.g., LUNA), applicable global and temporal summaries are used. The encoder remains frozen in the primary layer--readout experiments. \rev{We additionally evaluate a separate parameter-efficient adaptation setting on N170, in which trainable LoRA adapters are inserted into all 12 CBraMod transformer blocks while the original backbone weights remain frozen. The adaptation protocol is described in Section~\ref{sec:experimental_setup}.} Linear probes are trained on fixed feature summaries, whereas learned attention readouts fit query parameters on downstream training data. Detailed layer-to-readout mappings are provided in Appendix~\ref{app:A}.

\subsection{Disentangling Readout Mismatch from Depth Utility}
\rev{To evaluate the predictive utility associated with layer selection under a controlled readout protocol, we employ a sequential selection procedure on development data. Let $\widehat{P}_{\mathrm{dev}}(\ell,r)$ denote the validation performance obtained using candidate-specific
regularization.} Within each outer fold, we first select the optimal final-layer readout $r_L$, and subsequently identify the optimal depth $\ell_{\rm depth}$ under that chosen readout:
\begin{equation}
 r_L = \arg\max_{r \in \mathcal{R}} \widehat{P}_{\rm dev}(L,r), \qquad
 \ell_{\rm depth} = \arg\max_{\ell} \widehat{P}_{\rm dev}(\ell, r_L).
 \label{eq:selection}
\end{equation}
We also determine the globally optimal depth-readout pair $(\ell_*, r_*) = \arg\max_{(\ell,r)} \widehat{P}_{\rm dev}(\ell,r)$. Evaluating these fitted candidates on held-out participants yields three core performance contrasts:
\begin{align}
 \Delta_{\rm readout} &= P_{\rm test}(L, r_L) - P_{\rm test}(L, r_{\rm mean}), \label{eq:readout} \\
 \Delta_{\rm depth} &= P_{\rm test}(\ell_{\rm depth}, r_L) - P_{\rm test}(L, r_L), \label{eq:depth} \\
 \Delta_{\rm joint} &= P_{\rm test}(\ell_*, r_*) - P_{\rm test}(L, r_L). \label{eq:joint}
\end{align}
Here, $\Delta_{\rm readout}$ quantifies the performance restored to the final layer by replacing naive global mean pooling ($r_{\rm mean}$) with an optimized readout. Our primary metric, $\Delta_{\rm depth}$, \rev{measures the additional predictive utility obtained by selecting a different layer while retaining the final-layer-selected readout} $r_L$. Feature standardization, regularization parameters, and probe heads are fitted independently at each layer to account for layerwise distributional shifts.

\subsection{\rev{Residual Analysis Across Feature Dimensions}}
\label{sec:conditional_methods}

\rev{
We quantify the incremental predictive utility of
intermediate-layer residuals after fitting a mapping
from final-layer to intermediate-layer features.
The primary analysis uses PCA200 representations,
while additional dimensionality controls examine
how residual utility changes with the feature
dimensions of both representations.
}

Within each outer fold, intermediate features $z_m$ (for $m=\ell_{\rm depth}$) and final-layer features $z_L$ are standardized using statistics estimated from the outer-training split and independently reduced to 200 dimensions by Principal Component Analysis (PCA). PCA is fitted only on the outer-training data and then applied unchanged to held-out participants, yielding $\tilde{z}_m,\tilde{z}_L\in\mathbb{R}^{200}$. We fit a linear mapping $\widehat{A}$ from final to intermediate features using ridge regression, selecting regularization hyperparameters via inner cross-validation. To prevent data leakage during probe training, we compute cross-fitted residuals across participant folds:
\begin{equation}
e^{\rm cf}_{m\mid L} = \tilde{z}_m - \widehat{A}_{-k}\tilde{z}_L,
\label{eq:residual}
\end{equation}
where $\widehat{A}_{-k}$ is estimated excluding participant fold $k$. Test residuals are computed using the full outer-training mapping $\widehat{A}_{\rm tr}$.

\rev{
To evaluate the sensitivity of residual augmentation to feature dimensionality, we repeat this procedure under different final-layer and intermediate-layer feature dimensions. Specifically, we consider $(d_L,d_m)\in\{(100,100),(200,200),(400,400),(400,200),(d_L^{\rm native},200),(d_L^{\rm native},d_m^{\rm native})\}$, where $d_L$ and $d_m$ denote the final-layer and intermediate-layer feature dimensions, respectively. We denote the corresponding final-layer features and cross-fitted residuals by $\tilde{z}_L^{(d_L)}$ and $e_{m\mid L}^{\rm cf,(d_L,d_m)}$. PCA is omitted only for the representation retained at its native dimension. The selected layer, readout, and participant splits are held fixed across configurations, with all fitted transformations estimated exclusively from outer-training participants.
}

\rev{
For each dimensionality configuration, we first quantify the absolute residual augmentation gain relative to the final-layer baseline:
}
\begin{equation*}
\rev{
\Delta_{\rm aug}(d_L,d_m)
=
P_{\rm test}\left(
[\tilde{z}_L^{(d_L)},e_{m\mid L}^{\rm cf,(d_L,d_m)}]
\right)
-
P_{\rm test}\left(\tilde{z}_L^{(d_L)}\right).
}
\end{equation*}

We also compare aligned residual augmentation against a trial-shuffled residual control:
\begin{equation}
\rev{
E_{\rm resid}(d_L,d_m)
=
P_{\rm test}\left(
[\tilde{z}_L^{(d_L)},e_{m\mid L}^{\rm cf,(d_L,d_m)}]
\right)
-
\frac{1}{S}\sum_{s=1}^{S}
P_{\rm test}\left(
[\tilde{z}_L^{(d_L)},\Pi_s e_{m\mid L}^{\rm cf,(d_L,d_m)}]
\right).
}
\label{eq:excess}
\end{equation}
\rev{
Here, $\Pi_s$ denotes a trialwise permutation operator applied independently within each participant in the training and test splits, with $S=20$ independent permutations. Each shuffled control is evaluated using a separately refitted augmented probe. Thus, $\Delta_{\rm aug}$ measures the overall gain from residual augmentation, whereas $E_{\rm resid}$ measures the additional predictive utility associated with trialwise alignment relative to the shuffled control. Both contrasts use the same feature dimensions and downstream evaluation protocol.
}

\subsection{Cross-dataset Transfer and Subspace Interventions}

To evaluate the transferability of layer depth and residual dynamics, we freeze the source-selected configuration $(\ell_s,r_s)$ and evaluate it on target datasets, re-fitting only downstream classification heads and standardizations using target training data. Layer controls use the same readout: Final, \rev{First (the first encoder block, H01),} the mean score over all non-final layers (Random), and a layer selected on target development participants (Target). \rev{The source-selected layer and readout are determined using source development data and held fixed during target evaluation, whereas Target selects the layer using target development participants under the same readout.}

To assess the transferability of source-derived predictive directions, we construct a task-relevant subspace from source residuals and evaluate both coordinate augmentation of the target final-layer features and direction removal from the target intermediate representation. Let $Q_s$ denote an orthonormal basis capturing the mean residual directions across task classes in the source dataset. For sleep staging, four task directions are obtained from class-mean source residuals in a source-fitted PCA200 space; their projected target residual coordinates are appended to the final-layer features. \rev{Specifically, for HMC-to-ISRUC transfer, source standardization, PCA, and the ridge mapping from final-layer to selected-layer features are fitted on HMC and applied unchanged to ISRUC. Participant-equal class-mean residuals form five sleep-stage prototypes, which are centered before their leading four right singular vectors are used to construct $Q_s$.}

\rev{Here $Q_s$ and the residuals share the selected-layer PCA200 coordinates. Deletion acts on $\tilde{z}_{m,s}$, the target intermediate features transformed by source-fitted standardization and PCA:}
\begin{equation}
\rev{\tilde{z}_{m,s}^{(-Q)}
=
\tilde{z}_{m,s}-\gamma Q_s Q_s^\top(\tilde{z}_{m,s}-\mu_{\rm target}),}
\label{eq:direction_removal}
\end{equation}
where $\mu_{\rm target}$ is the target training mean \rev{in these source-fitted PCA coordinates} and $\gamma=1$ corresponds to complete subspace removal. We evaluate performance drops relative to rank-matched controls (random orthogonal subspaces, residual PCA components, and label-shuffled controls) at $\gamma=1$, averaging 20 independently evaluated random and shuffled bases (Appendix~\ref{app:strict_frozen}). Finally, we measure functional recovery by re-fitting downstream classification heads and target standardization on the modified representations, distinguishing frozen-head dependence from predictive utility remaining after adaptation.

\subsection{Experimental Setup and Evaluation \rev{Protocols}}
\label{sec:experimental_setup}
We evaluate five representative EEG foundation models (CBraMod, REVE, LaBraM, LUNA, and CSBrain) on motor imagery (OpenBMI; \citealp{lee2019openbmi}), two ERP tasks (ERP CORE N170 and N400; \citealp{kappenman2021erpcore}), and emotion recognition (SEED-V; \citealp{liu2021seedv}). Within-task cross-dataset transfer uses HMC \citep{alvarez2021hmc} to ISRUC-1 \citep{khalighi2016isruc} for sleep staging, and OpenBMI to BCI Competition IV-2a \citep{brunner2008bcic} for motor imagery. \rev{Additional pairs are OpenBMI--WBCIC-SHU, Benchmark--BETA (SSVEP), and ERP CORE P3--Won2022 (P300).}

Evaluation strictly follows participant-disjoint outer folds to prevent selection bias \citep{varma2006bias,cawley2010selection}. Performance is measured using participant-balanced accuracy (BA) for motor imagery and SEED-V, participant-balanced AUROC for ERP paradigms, and macro-F1 for sleep staging. \rev{For signal-derived physiological probing, we additionally report the participant-averaged coefficient of determination ($R^2$).} Statistical significance is assessed via paired 95\% bootstrap confidence intervals resampled at the participant level across all conditions. All implementation and pre-processing details are provided in Appendices~\ref{app:A}--\rev{\ref{app:additional}}.

\rev{
For signal-derived probing, we compare raw EEG, three independently initialized CBraMod encoders, and all 12 layers of pretrained CBraMod using Ridge regression. PCA200 is used for the primary comparison, with native-dimensional controls for raw EEG, H01, and H12. For parameter-efficient adaptation, rank-8 LoRA adapters ($\alpha=16$) are applied to all 12 CBraMod blocks on N170, while the original backbone weights remain frozen. Both experiments use five outer participant folds and three seeds; seed-level scores are averaged within participants before bootstrap estimation. Adapted representations are evaluated using freshly fitted native-dimensional diagnostic probes, separately from the adaptation classifier. We evaluate both the pretrained-validation-selected layer after adaptation and a layer independently reselected using inner validation, without accessing outer-test labels.
}

%% file: sections/04_results.tex
\section{Results}
\setlength{\textfloatsep}{12pt plus 2pt minus 2pt}

\subsection{Depth utility after final-layer readout selection}

\begin{figure}[!htp]
\setlength{\abovecaptionskip}{6pt}
\centering
\includegraphics[width=0.95\linewidth]{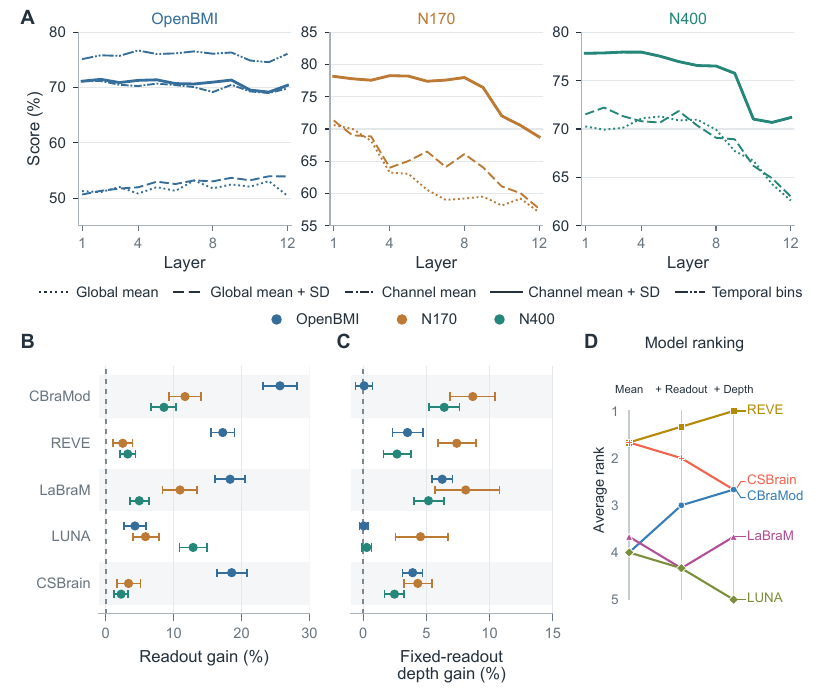}
\caption{\rev{\textbf{Different evaluation interfaces reshape apparent model quality.} \textbf{A:} CBraMod layer profiles under five readouts. \textbf{B--C:} Final-layer readout gains and additional fixed-readout depth gains; task colors are shared with A. \textbf{D:} Average ranks across the three primary tasks under mean-pooled final-layer evaluation, final-layer readout selection, and fixed-readout depth selection.}}
\label{fig:depth}
\end{figure}

We first separate readout mismatch from depth effects (H1). Selecting the readout at the final layer improves prediction over global mean pooling in all 15 model--task conditions, with positive paired confidence intervals (Figure~\ref{fig:depth}B; Table~\ref{tab:full}). The effect is particularly large on OpenBMI for CBraMod and LaBraM, showing that weak mean-pooled final-layer performance can substantially underestimate the information available at the final layer.

After fixing this final-layer-selected readout, choosing the layer provides an additional improvement in 12 of 15 conditions (Figure~\ref{fig:depth}C). The size of this depth benefit varies considerably across models and tasks. In particular, models with similarly weak mean-pooled final-layer performance can show very different \rev{additional depth gains} from moving to an intermediate layer. Readout mismatch and depth utility therefore capture distinct effects.

Joint layer-readout selection chooses the same readout as final-layer selection in every \rev{primary model--task} condition, giving $\uinterface=\udepth$. The selected readout is also stable across all five outer folds within each condition, whereas the preferred layer can vary (Table~\ref{tab:stability}). Several conditions show little additional depth gain, indicating that intermediate layers are not uniformly preferable once readout mismatch has been accounted for.

\rev{These interface choices also alter model rankings. Across the three primary tasks, REVE's average rank improves from 1.67 to 1.00, while CBraMod improves from 4.00 to 2.67 (Figure~\ref{fig:depth}D).} \rev{The selected layer also outperforms the final layer under both learned readout families (Appendix~\ref{app:B}, Figure~\ref{fig:conditional}; Table~\ref{tab:learned}).} \rev{Increasing fitting labels from 10\% to 100\% raises OpenBMI readout gains by 8.43--8.82\%, compared with 0.38--1.91\% for joint-selection gains, with the validation pool fixed (Appendix~\ref{app:B}, Figure~\ref{fig:label_budget}; Table~\ref{tab:budget}). More fitting labels therefore do not remove the importance of readout choice.} Full layer-wise curves appear in Figure~\ref{fig:oracle_envelope}.

\subsection{\rev{Residual predictive utility depends on feature dimensionality}}

\rev{We next examine whether intermediate-layer residuals
provide additional predictive utility under a compressed
feature interface.} In the PCA200 analysis, cross-fitted residual augmentation improves final-layer prediction in 13 of 15 conditions. Relative to shuffled residuals matched in distribution, covariance, and scale, the residual contribution $E_{\rm resid}$ also has a positive paired confidence interval in 13 of 15 conditions (Figure~\ref{fig:residual_extra}A; Table~\ref{tab:crossfit}). \rev{Nonlinear mappings and task heads also retain positive residual gains in the diagnostic conditions (Appendix~\ref{app:capacity}).}

\rev{The utility of residual representations depends strongly on dimensionality. In our dimensionality scan, expanding the final-layer representation from PCA200 to its native dimension while keeping the residual features at PCA200 reduces augmentation gains for CBraMod and REVE from 9.57\% and 8.96\% to 1.22\% and 0.48\% on N170, and from 6.11\% and 2.12\% to 0.71\% and $-0.01\%$ on N400. When both representations are kept at their native dimensions, these differences drop to $-0.16\%$ and $+0.25\%$ on N170, and $+0.14\%$ and $-0.99\%$ on N400, showing no consistent improvement over the baseline (Table~\ref{tab:dimension}). Likewise, under a matched 200-feature budget, neither residual augmentation nor fusion outperforms the stronger single-layer baseline. Together, these findings indicate that the incremental predictive value of residual features is interface-dependent, rather than reflecting information intrinsically absent from the full final-layer representation. Detailed controls for dimensionality, model capacity, and fusion strategies are provided in Appendix~\ref{app:capacity}.}

\subsection{Reusing source-selected \rev{layer--readout interfaces} across datasets}

\begin{figure}[!htp]
\setlength{\abovecaptionskip}{6pt}
\centering
\includegraphics[width=0.85\linewidth]{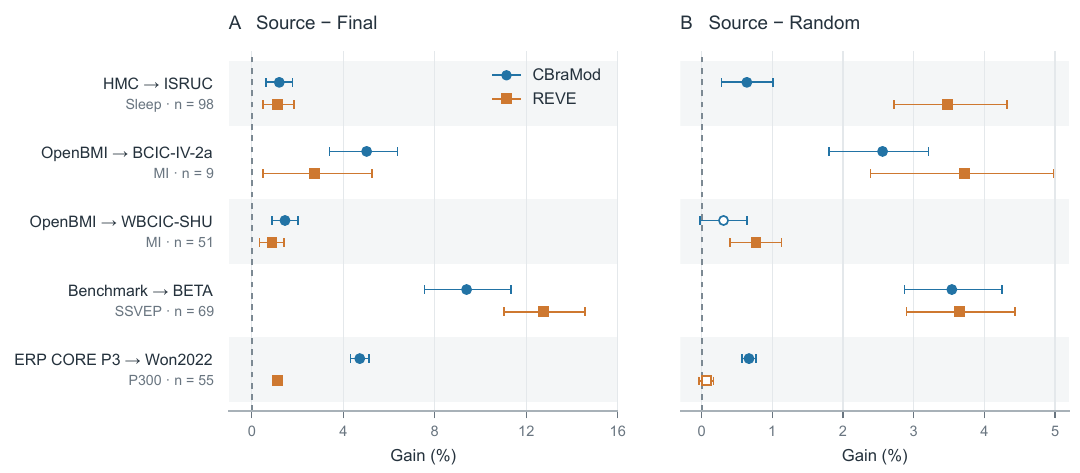}
\caption{\rev{\textbf{Source-selected depth transfers across datasets.} \textbf{A:} Source-selected layer probing minus Final layer probing. \textbf{B:} Source-selected layer probing minus random-layer expectation.}}
\label{fig:transfer}
\end{figure}

We next test whether \rev{layer--readout interfaces} selected on one dataset remain useful on another. \rev{Across five
dataset pairs and four task families, source-selected
layers outperform Final in all ten CBraMod/REVE
comparisons. Gains over Random have positive intervals
in eight comparisons (Figure~\ref{fig:transfer}).}

\rev{
Table~\ref{tab:transfer} further illustrates these transfer results
through absolute performance and additional
layer-selection controls. Source-selected interfaces
closely approach target-validation-selected
performance in sleep staging and motor imagery.
Importantly, source selection also outperforms the
first-layer baseline in seven of ten CBraMod/REVE
comparisons, indicating that the transferred gains
are not explained solely by a fixed preference for
shallow representations. The improvements are
particularly pronounced for REVE on HMC-to-ISRUC
and OpenBMI-to-BCIC-IV-2a, where source selection
exceeds First by 12.96 and 8.60 \%, respectively. For MI transfer, CBraMod selects First (H01), whereas REVE selects H11 and outperforms First by 8.60\%, showing that source selection is not simply an early-layer heuristic.
}

\rev{We additionally examine whether selected depths transfer across task families. OpenBMI-selected layers exceed random-layer averages by 1.73 and 1.89 \% on N170 and N400, respectively, while showing similar performance to the first layer. In the reverse N170-to-OpenBMI direction, source selection improves over First by 0.69 \%. These results suggest that depth preferences can also transfer across tasks, with
the magnitude of the benefit depending on the target paradigm (Appendix~\ref{app:transfer_extra}).
}

\input{tables/table_reuse}

\subsection{Source residual directions remain useful after transfer}

\begin{figure}[!t]
\setlength{\abovecaptionskip}{6pt}
\centering
\includegraphics[width=0.95\linewidth]{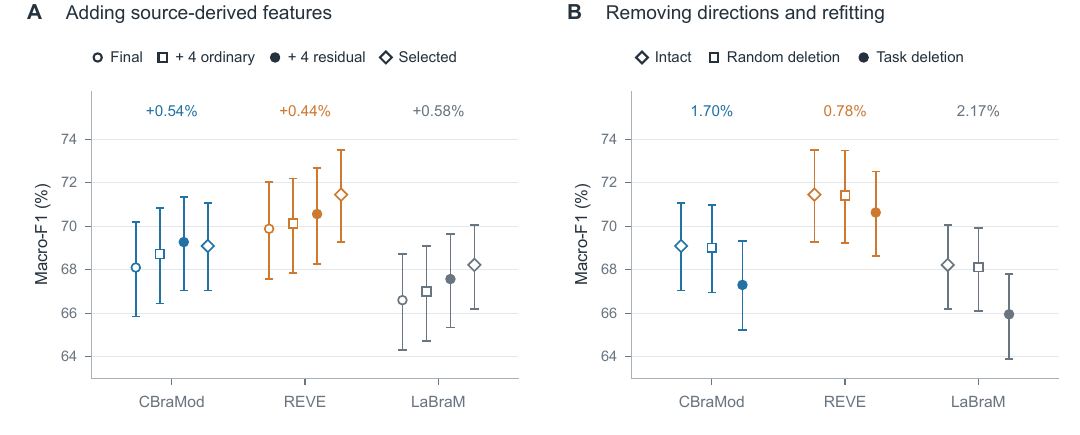}

\caption{\rev{\textbf{Source-defined task directions retain predictive utility after transfer.} In task transfer: HMC to ISRUC sleep staging ($n=98$). \textbf{A:} Augmentation of final-layer features with four source-derived residual coordinates, compared with ordinary supervised coordinates. \textbf{B:} Performance after removing source-defined versus random directions and refitting the target classifier. Labels indicate paired gains or excess losses; whiskers denote 95\% participant-bootstrap intervals. All analyses use source-fitted PCA200 features.}}
\label{fig:frozen}
\end{figure}

We next \rev{test whether a compact subspace defined from source residuals retains predictive utility on an unseen target dataset}. Source residuals are used to define a low-dimensional task subspace, which is then applied to target representations. \rev{Appending four source-derived residual coordinates to final-layer features improves target macro-F1 by 1.17, 0.67, and 0.97\% for CBraMod, REVE, and LaBraM (Figure~\ref{fig:frozen}A). Gains over ordinary supervised-coordinate augmentation, using the same four-dimensional feature budget, are 0.54, 0.44, and 0.58\%, all with positive paired intervals.}

Removing these directions with the target scaler and classifier frozen produces a pronounced performance drop \rev{for all three models (Appendix~\ref{app:strict_frozen}, Figure~\ref{fig:frozen_diagnostic}).} When the scaler and classifier are refit after removal, performance largely recovers, \rev{but deletion of the source-defined directions still incurs an additional loss of 1.70, 0.78, and 2.17\% relative to equal-rank random deletion (Figure~\ref{fig:frozen}B; Tables~\ref{tab:v5group2}--\ref{tab:v5controls}). Together, augmentation and post-refit deletion support transferable predictive utility beyond equal-dimensional controls in sleep staging. Frozen-head degradation reflects the target-fitted classifier's dependence on these directions, while alternative predictive structure remains available after refitting.}

\subsection{\rev{Signal-derived Feature Accessibility and Adaptation}}

\rev{
To clarify why intermediate layers excel when frozen, we first probe signal-derived measurements across depth using PCA200 Ridge probes, with additional native-dimensional controls for raw EEG, H01, and H12 (Appendix~\ref{app:training_extra}). Pretraining affects the accessibility of linear and nonlinear signal-derived measurements differently: in native features, it preserves time-domain N170 PO8 amplitude (raw ceiling $R^2=0.999$) at H01 ($0.876$ vs.\ $0.574$ random) before attenuating it at H12 ($0.181$ vs.\ $0.462$). Conversely, nonlinear OpenBMI band power---where raw-EEG Ridge fails, saturating regularization in every fold---becomes linearly decodable across the pretrained hierarchy ($R^2=0.665/0.486$ at H01/H12 vs.\ $-0.448/-0.674$ random). Thus, depth-wise gains track the linear accessibility of signal-derived measurements rather than high-level physiological representations.
}

\rev{
This depth profile generalizes across pretext objectives, yet remains contingent on keeping the encoder frozen. Across matched MAE and JEPA encoders, pretraining consistently lifts selected-layer N170 AUROC ($74.42\%$ and $74.47\%$ vs.\ $70.47\%$ random) even as final-layer specialization drops performance below random initialization ($69.42\%$ and $63.82\%$ vs.\ $70.59\%$). Once adapted via all-block LoRA, however, final-layer predictive performance improves: AUROC increases from $68.77\%$ to $79.57\%$, surpassing the fixed intermediate layer ($78.59\%$) and matching validation-based reselection ($79.54\%$; Appendix~\ref{app:adaptation}). Thus, the intermediate-layer advantage observed under frozen extraction is contingent on the adaptation setting and can be substantially reduced through parameter-efficient adaptation.
}

%% file: tables/table_reuse.tex
\begin{table}[!htp]
\centering
\caption{Absolute transfer scores (\%). Sleep: HMC to ISRUC and MI: OpenBMI to BCIC-IV-2a. All strategies use the same source
readout and refit the target classifier.
\rev{First denotes the first encoder block.}
Random averages non-final layers; Target selects
depth on target development participants.}
\label{tab:transfer}
\begin{tabular}{@{}llccccc@{}}
\toprule
Task & Model & Final & \rev{First} & Random & Source & Target \\
\midrule
Sleep & CBraMod & 70.56 & \rev{70.97} & 71.12 & 71.76 & 71.56 \\
Sleep & REVE    & 71.66 & \rev{59.83} & 69.30 & 72.79 & 72.97 \\
Sleep & LaBraM  & 69.22 & \rev{67.14} & 69.65 & 70.29 & 70.27 \\
MI    & CBraMod & 66.98 & \rev{71.99} & 69.43 & 71.99 & 71.72 \\
MI    & REVE    & 77.08 & \rev{71.22} & 76.10 & 79.82 & 79.82 \\
\bottomrule
\end{tabular}
\end{table}

%% file: sections/05_discussion.tex
\section{Discussion}
Our findings show that the apparent underperformance of final-layer EEG representations stems from two distinct factors. First, much of this deficit disappears when the final layer is paired with an appropriate readout. \rev{Second, intermediate layers frequently offer higher predictive utility even with the readout held fixed, though this advantage is tightly coupled to the feature interface: residual gains diminish sharply as final-layer dimensionality increases, yielding no reliable improvement once both representations retain their native dimensions. Together, these results indicate that readout design and layer selection govern complementary facets of representation accessibility---how effectively task information is extracted within a layer, and where along the encoder hierarchy it is most readily decodable.}

This distinction directly affects how frozen foundation models should be benchmarked. A single pooling rule confounds layer depth with readout mismatch, making layer comparisons meaningful only after an appropriate readout is established. Even under matched readouts, intermediate layers improve performance across most conditions and substantially shift model rankings, lifting average rank across the three tasks from 1.67 to 1.00 for REVE and from 4.00 to 2.67 for CBraMod.
\rev{Crucially, this depth preference is specific to frozen extraction. Under all-block LoRA adaptation on N170, the final layer reaches 79.57\% AUROC---surpassing the fixed intermediate layer (78.59\%) and matching validation reselection (79.54\%). Layer utility therefore depends on the adaptation regime: depth selection is a primary lever when probing frozen representations, whereas parameter-efficient adaptation realigns task-relevant information with the model output.}

Beyond single-dataset probing, our transfer experiments show that useful depth structure generalizes across datasets.
\rev{
Across five dataset pairs and four task families, source-selected layer--readout interfaces outperform Final in all ten CBraMod/REVE comparisons, Random in eight, and First with positive paired confidence intervals in seven, showing that source-side selection identifies transferable interfaces beyond default final-layer extraction or heuristic layer choices.
}
Intervention analyses sharpen this geometric picture: a compact set of task-relevant directions from source residuals transfers directly to the target representation, where ablating them degrades a frozen classifier and appending them to final-layer features improves prediction.
\rev{
Although refitting the target classifier recovers most of this drop, ablating source-derived directions still incurs greater loss than equal-rank random deletion, confirming that residual subspaces capture reusable predictive coordinates in sleep staging rather than arbitrary variance.
}

Collectively, these findings argue against treating the final hidden state as the canonical representation of an EEG foundation model. Instead, depth, readout, and downstream adaptation jointly govern which predictive structures are accessible. Rigorous frozen-model benchmarking therefore requires reporting not only final-layer scores, but also the readout interface
\rev{and the task-dependent predictive utility distributed across depth}.

\textbf{Limitations.} The current experiments use lightweight mappings and focus the intervention analysis on a subset of the transfer settings. \rev{Residual analysis depends on the tested feature dimensions and mapping/probe families; it does not establish information-theoretic absence from final-layer representations. On SEED-V, readout helps some models, but depth gains remain inconclusive under video-level probability averaging. Our all-block LoRA experiment is limited to one model (CBraMod) and one dataset (N170).}

%% file: sections/06_conclusion.tex
\section{Conclusion}
\rev{
We show that intermediate-layer advantages in frozen EEG foundation models depend on both readout design and downstream task.
} Better readouts recover substantial performance at the final layer, while selecting depth at a fixed readout provides additional predictive utility in most conditions. These depth-dependent features can also remain useful across datasets, with source-derived predictive directions retaining value after transfer in sleep staging.
\rev{
These effects depend on representation dimensionality and adaptation: residual gains diminish with native-dimensional features, while the intermediate-layer advantage does not persist under the tested all-block LoRA setting. Overall, predictive utility varies across depth and depends on how representations are read out, transferred, and adapted.
} Evaluating EEG foundation models should therefore consider not only which layer performs best, but also how representations are read out and reused across datasets and downstream adaptation settings.

%% file: sections/supplementary.tex
\input{tables/supp_table_style}
\raggedbottom
\setlength{\intextsep}{12pt plus 2pt minus 2pt}
\section{Experimental details}
\label{app:A}

\subsection{Datasets, readouts, and preprocessing}
All data splits are participant-disjoint, and all sessions from the same participant remain in the same split. ERP CORE N170 and N400 share the same participant cohort. SEED-V probabilities are averaged within video before participant-equal scoring. For HMC-to-ISRUC transfer, we use the 103-participant HMC development subset as the source and 98 ISRUC participants as the target after excluding S008 and S040 for channel/reference incompatibility.

Readouts operate directly on frozen encoder tokens. For $H\in\mathbb R^{C\times T\times d}$, global mean averages channels and time, while global mean+std additionally summarizes token dispersion. Channel mean averages over time while preserving channel identity, and channel mean+std additionally retains temporal dispersion within each channel. The structure readout concatenates channel summaries from four fixed temporal bins when the tokenization permits this. LUNA has no physical channel axis and therefore uses the applicable global and temporal summaries. Each readout is standardized using training-fold statistics before fitting the downstream classifier.

The sleep interface uses four mastoid-referenced leads. Each 30-s epoch is represented by six 5-s encoder outputs, with temporal dispersion computed within segment before segment averaging.

\subsection{Model checkpoints and evaluation protocol}
We evaluate CBraMod, REVE-base, LaBraM-base, LUNA-base, and CSBrain with frozen encoders \rev{in the primary experiments.} REVE's released pretraining inventory includes OpenBMI.

All layer--readout candidates use the same outer and development folds. Regularization is selected on development data for each candidate, and standardization and dimensionality reduction are fitted only on the corresponding training data. Participant-level contrasts use paired participant bootstrap intervals. ISRUC macro-F1 is computed from pooled participant confusion counts within each bootstrap resample.

\rev{Seeds are averaged within participant before bootstrap resampling. Intervals are unadjusted for multiple comparisons. The transfer experiments use five dataset pairs: HMC--ISRUC (sleep, 98 target participants), OpenBMI--BCIC-IV-2a (MI, 9), OpenBMI--WBCIC-SHU (MI, 51), Benchmark--BETA (SSVEP, 69), and ERP CORE P3--Won2022 (P300, 55).}

\clearpage
\section{\rev{Readout, depth, and model comparison}}
\label{app:B}
\subsection{\rev{Absolute scores, rankings, and paired gains}}
\label{app:rankings}
\rev{Table~\ref{tab:depth} combines absolute scores and model rankings under the three evaluation protocols. Table~\ref{tab:full} gives paired gains and intervals, including SEED-V. Average ranks across OpenBMI, N170, and N400 are shown in the main Discussion.}
\input{tables/b_absolute}
\input{tables/b_full}

\rev{On SEED-V, all five fixed-readout depth intervals include zero, while readout selection helps CBraMod and LaBraM. Evaluation averages window probabilities within video and then weights participants equally; it does not fit a video-level temporal model.}

\subsection{\rev{Learned readouts, label budget, and selection stability}}
\rev{The global-query readout learns one query over all tokens. The channel-preserving readout learns a shared temporal query within each channel and concatenates channel features. Both use frozen tokens and participant-disjoint folds, fitting parameters at each layer and averaging three seeds. N170 has one token per channel, so channel-preserving attention has unit weight.}
\input{tables/b_learned}

\begin{figure}[H]
\centering
\includegraphics[width=\linewidth]{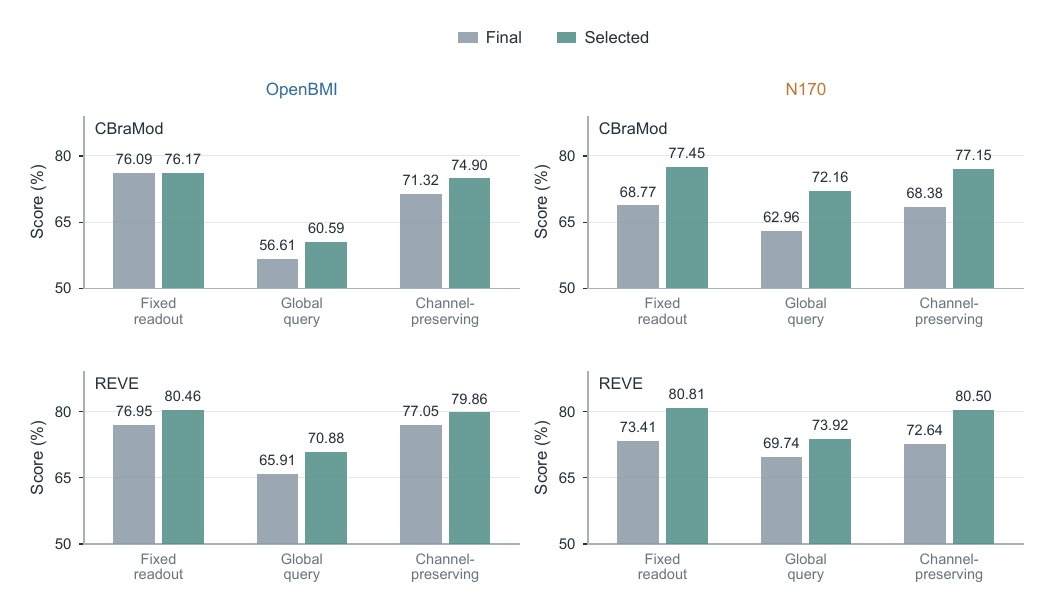}
\caption{\rev{\textbf{Depth gains persist with learned readouts.} Final and selected-layer scores for CBraMod and REVE on OpenBMI (BA) and N170 (AUROC). Learned readouts average three seeds.}}
\label{fig:conditional}
\end{figure}

The reduced-label analysis uses 10\%, 25\%, 50\%, and 100\% of the outer-training data, with stratified subsampling by participant and class while keeping validation and test participants fixed. Table~\ref{tab:budget} reports the resulting \rev{readout and joint-interface gain changes.}
\input{tables/b_budget}

\begin{figure}[H]
\centering
\includegraphics[width=\linewidth]{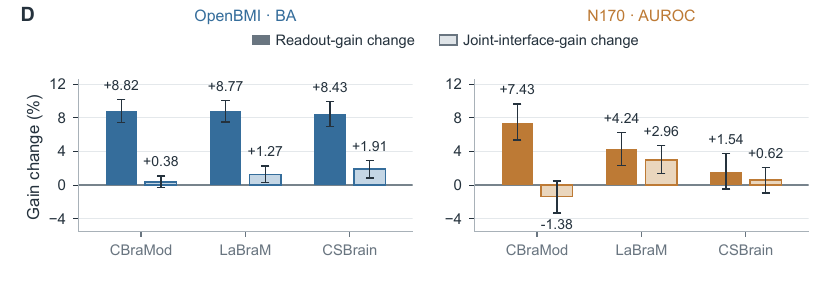}
\caption{\rev{\textbf{Readout choice remains important as fitting labels increase.} Changes in readout and joint layer--readout gains from 10\% to 100\% fitting labels, with the validation pool fixed. Light bars show joint-interface gain changes. OpenBMI uses BA; N170 uses AUROC. Intervals are paired 95\% participant-bootstrap intervals.}}
\label{fig:label_budget}
\end{figure}

Readout selection is stable across the five outer folds for OpenBMI, N170, and N400, whereas the selected layer can vary across folds (Table~\ref{tab:stability}). SEED-V shows greater variation in the selected readout.
\input{tables/b_stability}

\clearpage
\subsection{\rev{Layer-wise readout profiles}}
\label{app:oracle}
\begin{figure}[H]
\centering
\includegraphics[width=\linewidth]{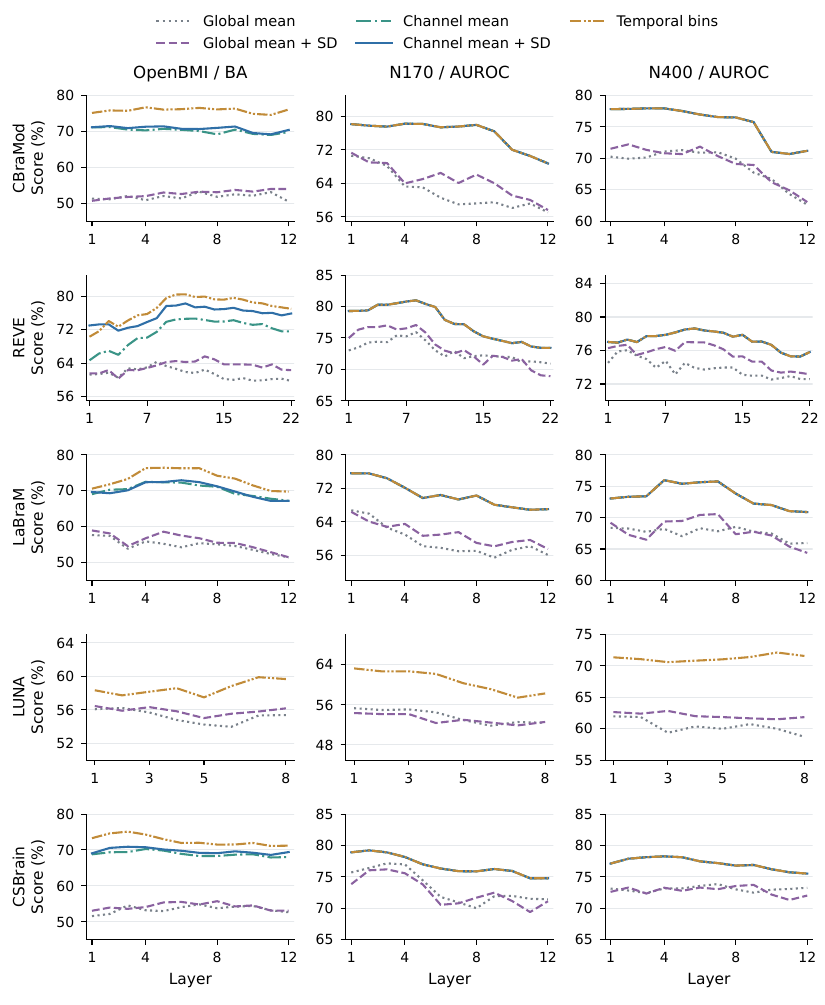}
\caption{\rev{Layer-wise performance under five readouts. Rows identify models and columns identify tasks; colors and line styles share one legend. Scores are participant-average BA (OpenBMI, $n=54$) or AUROC (N170/N400, $n=40$). LUNA uses the three readouts applicable to its token layout.}}
\label{fig:oracle_envelope}
\end{figure}

\rev{Figure~\ref{fig:oracle_envelope} shows every evaluated layer and readout for the three primary tasks. These descriptive curves complement the development-selected comparisons above.}

\clearpage
\section{\rev{Residual utility and decoder controls}}
\label{app:C}\label{app:capacity}
\subsection{Cross-fitted residual analysis}
\label{app:crossfit}
Intermediate and final representations are standardized and independently reduced to 200 principal components using transformations fitted on the outer-training fold. Ridge regularization for the final-to-intermediate map is selected on participant-grouped inner splits. Training residuals are then generated by participant-grouped cross-fitting, while outer-test residuals use a map refitted on the full outer-training data.

The shuffled control permutes residual rows within participant, separately in training and test data, and refits the augmented probe for each of 20 permutations. This preserves the within-participant residual distribution while breaking trial correspondence with the final-layer features. \rev{Tables~\ref{tab:crossfit}--\ref{tab:crossfitchecks} pair augmentation gains with absolute scores under the same balanced logistic probe.}
\input{tables/f_crossfit}
\input{tables/f_crossfit_checks}

\clearpage
\subsection{\rev{Representation dimension}}
\rev{The cross-fitted scan fixes each fold's layer and readout while varying final/selected feature counts over 100/100, 200/200, 400/400, 400/200, full/200, and full/full. Standardization and PCA use training participants only. Keeping 200-dimensional intermediate residuals while expanding final features to full size tests the effect of final-layer compression.}
\begin{figure}[H]
\centering
\includegraphics[width=0.90\linewidth]{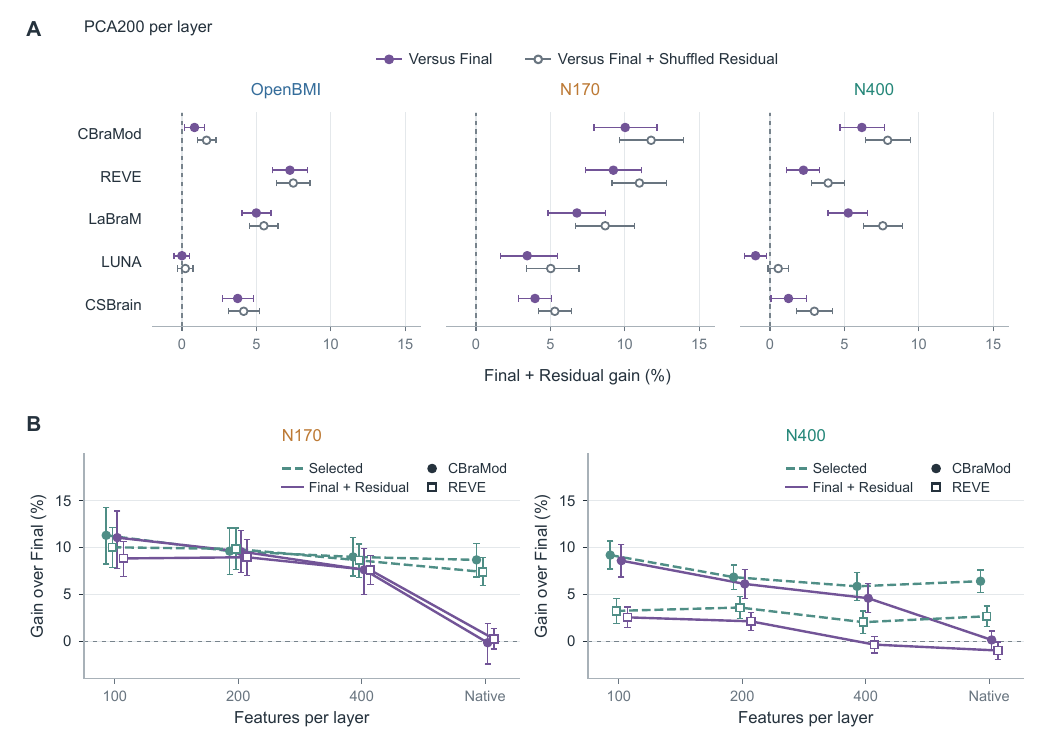}
\caption{\rev{Residual utility across models and feature dimensions. \textbf{A:} PCA200 cross-fitted gains over Final (filled) and shuffled augmentation (open). \textbf{B:} Fixed-layer/readout dimension scan: Selected minus Final and Final + residual minus Final. Native denotes full model-specific dimensions. Intervals are paired participant-bootstrap 95\% CIs; the two panels use their respective fitted protocols.}}
\label{fig:residual_extra}
\end{figure}
\rev{Table~\ref{tab:v5sensitivity} retains the original unreduced ERP and Gaussian-projection comparisons. All four unreduced augmentation estimates are negative. Together with the cross-fitted scan, they show the sensitivity of augmentation to feature dimension and high-dimensional estimation.}
\input{tables/f_sensitivity}

\clearpage
\input{tables/extra_dimensions}

\subsection{\rev{Matched-budget multilayer fusion}}
\rev{Compact methods use 200 total classifier-input features on the same participant-disjoint folds. Three-layer methods use first, middle, and final layers. Scalar fusion uses a common training-fitted PCA basis and three learned softmax weights. Uncompressed concatenation is shown separately because it uses more features than Selected. Residual augmentation exceeds Selected only for CBraMod/OpenBMI by 0.33\% [$-0.67$, 1.34], where Final is still stronger.}
\input{tables/extra_fusion}

\clearpage
\subsection{\rev{Mapping and task-head capacity}}
\rev{Capacity controls compare ridge, a one-hidden-layer MLP of width 256, and a two-hidden-layer MLP of width 512 with dropout 0.25 and weight decay 0.01. Task heads use balanced logistic regression or a width-128 MLP. Each comparison uses the same head family for Final, augmented, and shuffled-augmented features. Wider nonlinear mappings retain positive predictive gains in the diagnostic conditions, but do not improve held-out reconstruction over ridge (Tables~\ref{tab:capacity}--\ref{tab:mapper}).}
\input{tables/supp_capacity}
\input{tables/extra_mapper}

\clearpage
\section{\rev{Cross-dataset and cross-task layer reuse}}
\label{app:D}\label{app:transfer_extra}
\subsection{\rev{Transfer settings and layer controls}}
\rev{Within each dataset pair, source-selected depth and the chosen readout are fixed for target evaluation; classifiers are fitted on target training labels. Random averages scores from independently fitted non-final layers, and Target selects depth on target development participants at the same readout. First and Middle provide fixed-layer controls.}

The OpenBMI-to-BCIC-IV-2a experiment uses the shared 21-channel montage Fz, FC3, FC1, FC2, FC4, C5, C3, C1, Cz, C2, C4, C6, CP3, CP1, CPz, CP2, CP4, P1, Pz, P2, and POz with a common-average reference. Features are re-extracted under this common input protocol. Channel mean plus temporal standard deviation is fixed as the source readout. Source development data select CBraMod H01 and REVE H11, which are then transferred unchanged to BCIC-IV-2a; H12 and H22 serve as their final-layer comparators. Target evaluation uses nine leave-one-participant-out outer folds, with the scaler, regularization, and classifier refitted within each target fold.

\rev{WBCIC-SHU and BETA use full channel mean plus standard deviation; Won2022 uses source-selected channel mean. The OpenBMI choices for BCIC-IV-2a and WBCIC-SHU come from separate source-selection runs. Tables~\ref{tab:selection_extra}--\ref{tab:transfer_ci} give fixed-layer controls and paired gains for the main transfer comparisons.}
{\setlength{\intextsep}{12pt plus 2pt minus 2pt}
\input{tables/extra_selection}
\input{tables/transfer_gains}
\input{tables/extra_transfer_ci}
}

\subsection{\rev{MI participant effects and classical baseline}}
\rev{The original REVE/BCIC-IV-2a gain is 2.28\% [0.42, 4.24], with positive effects in 7/9 participants and an exact two-sided sign-flip $p=0.0781$. Table~\ref{tab:mi_followup} retains these participant effects. This test uses the initial Final score of 77.55\%; the matched refit in Table~\ref{tab:transfer} gives 77.08\%, with Source unchanged at 79.82\%.}
\input{tables/d_mi_followup}
For participant-level depth effects $d_i$, the exhaustive two-sided sign-flip statistic is
\begin{equation}
 p_{\rm sf}=2^{-9}\sum_{s\in\{-1,1\}^9}
 \mathbb{I}\!\left\{\left|\sum_{i=1}^{9}s_i d_i\right|\geq
 \left|\sum_{i=1}^{9}d_i\right|\right\}.
 \label{eq:signflip}
\end{equation}
This calculation treats participant-effect signs as exchangeable under the null and is reported alongside the paired bootstrap intervals.

\rev{CSP uses the same accepted left/right trials (2,592), 21 channels, and participant splits. Inner development participants select among 8--30, 8--12, and 13--30 Hz bands and 2/4/6/8 components; covariance uses Ledoit--Wolf estimation and LDA uses automatic shrinkage. CSP BA is 62.62\%, compared with 66.98\% for CBraMod Final and 71.99\% for H01. H01 exceeds CSP by 9.38\% [2.93, 15.47], while its source-validation score nearly ties H03 (71.18 versus 71.14\%).}

\subsection{\rev{Cross-task layer reuse}}
\rev{Source depth is the mode of development-selected layers, with ties toward shallower layers. The target's validation-selected final-layer readout is fixed and classifiers are refitted, so only layer identity transfers. N170 and N400 share the same 40-participant ERP cohort; OpenBMI has 54 participants.}
\input{tables/extra_crosstask}

\section{\rev{Source-direction transfer}}
\label{app:E}
\subsection{Source-residual intervention}
\label{app:strict_frozen}
For HMC-to-ISRUC transfer, the selected/final layers are H06/H12 for CBraMod and H14/H22 for REVE, \rev{and H07/H12 for LaBraM,} using channel means and temporal standard deviations. Source standardization and 200-dimensional PCA are fitted on HMC and applied unchanged to ISRUC.

\rev{A ridge map from final to selected-layer features is fitted on HMC, with regularization selected by participant-grouped validation. Participant-equal class means of source residuals form five sleep-stage prototypes. Centering them and taking their right singular vectors gives four task directions.}

\rev{Augmentation appends four projected target residual coordinates to the 200 final-layer features. Random, residual-PCA, shuffled-label, and ordinary supervised directions use the same coordinate count and fitting procedure. Ordinary directions come from the unresidualized source class means; both bases are also compared on the same residual representation.}

\rev{Deletion removes four directions from the selected-layer features in source-fitted PCA200 coordinates at unit strength (Equation~\ref{eq:direction_removal}). Random and within-participant label-shuffled controls average 20 independent bases. The frozen condition holds the target scaler and classifier fixed; refitting updates both, including regularization. Tables~\ref{tab:v5group2}--\ref{tab:v5controls} report scores and paired whole-participant bootstrap intervals, recomputing macro-F1 from pooled confusion counts.}
\input{tables/f_group2}
\input{tables/f_group2_controls}

\begin{figure}[H]
\centering
\includegraphics[width=\linewidth]{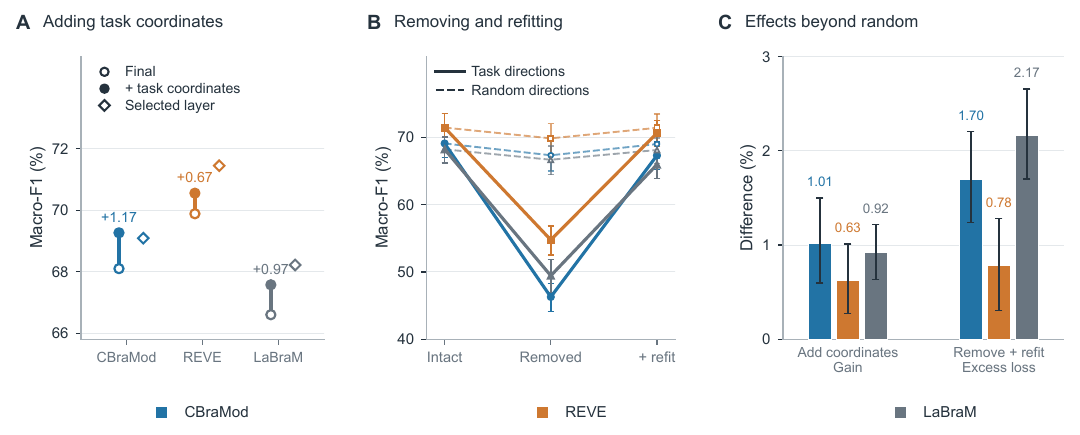}
\caption{\rev{\textbf{Complete source-direction transfer diagnostics.} HMC--ISRUC sleep staging ($n=98$), using source-fitted PCA200. \textbf{A:} Final, source-residual augmentation, and Selected macro-F1. \textbf{B:} Intact, rank-four deletion with the target scaler and classifier frozen, and refitted scores for task and random directions. \textbf{C:} Augmentation gains and post-refit excess deletion losses relative to random controls. Random controls average 20 bases; source transformations remain fixed. Intervals are 95\% whole-participant bootstrap intervals.}}
\label{fig:frozen_diagnostic}
\end{figure}

\clearpage
\subsection{\rev{Energy-matched perturbations}}
\rev{Energy matching rescales control projections to match displacement in target-standardized coordinates. It is distinct from complete rank-four deletion and is reported separately in Table~\ref{tab:energy_extra}. The positive augmentation and post-refit direction-transfer evidence applies to sleep staging; the exploratory MI residual experiment does not reproduce this pattern.}
\input{tables/extra_energy}



\clearpage
\section{\rev{Physiology, adaptation, and pretraining}}
\label{app:F}\label{app:additional}
\subsection{\rev{Physiological measurements}}
\label{app:training_extra}
\rev{Signal-derived probing compares raw EEG, three independently initialized CBraMod encoders (seeds 1--3), and all twelve pretrained CBraMod layers on the same accepted trials and preprocessing. Targets are log$_{10}$ mean C3/C4 Welch power over 8--30 Hz and baseline-corrected PO8 mean amplitude over 150--200 ms. Readouts are channel mean for N170 and structure for OpenBMI. PCA200 is the primary comparison; native controls use raw EEG, H01, and H12 (Table~\ref{tab:physiology}).}

\rev{Evaluation uses five participant-disjoint outer folds and three inner folds. Ridge penalties $\{10^{-4},10^{-3},10^{-2},0.1,1,10,100\}$ maximize participant-mean inner-validation $R^2$. Standardization and PCA are fitted only on training participants. Random-initialization scores are averaged within each held-out participant before 10,000 paired participant-bootstrap resamples; seeds do not increase the participant count.}
\input{tables/extra_physiology}

\clearpage
\subsection{\rev{All-block LoRA adaptation}}
\label{app:adaptation}
\rev{On N170, rank-8 LoRA adapters ($\alpha=16$) are inserted into spatial/temporal attention QKV and output projections and both feedforward matrices in all twelve CBraMod blocks (307,200 trainable encoder parameters). Original weights, normalization parameters, and patch embedding remain frozen. AdamW uses adapter/head learning rates $10^{-4}/10^{-3}$, weight decay 0.01, batch size 64, and gradient clipping at 1.}

\rev{Five outer folds and three seeds each use three independently trained inner encoders, excluding validation participants from gradient updates. Inner validation selects training duration (maximum 30 epochs, patience five); the outer encoder trains for the median inner duration, giving 60 fitted checkpoints. Adapted native channel-mean features receive freshly fitted balanced L2 logistic probes with $C\in\{10^{-5},10^{-4},\ldots,10^3\}$. Layer and $C$ selection use inner-checkpoint validation features only. Fixed selected retains the pretrained-validation-selected layer; Reselected searches all twelve adapted blocks. Table~\ref{tab:adaptation} separates diagnostic probes from the adaptation classifier and reports all three seeds. Seed scores are averaged within participant before 10,000 paired bootstrap resamples.}
\input{tables/extra_adaptation}

\clearpage
\subsection{\rev{Controlled pretraining}}
\label{app:pretraining}
\rev{MAE and JEPA share an eight-block encoder, tokenizer, 25-channel HBN input, sampling stream, temporal masking, optimizer schedule, and nominal exposure. MAE reconstructs masked waveform patches; JEPA predicts normalized teacher representations. Three paired seeds train for 50,000 updates with global batch 256 (microbatch 32, accumulation eight). Teacher computation and output heads differ, so this compares pretraining schemes.}

\rev{Pretraining excludes windows with RMS above 100 $\mu$V or maximum channel peak-to-peak amplitude above 500 $\mu$V, retaining 12,792 windows (17.77 hours). The downstream comparison includes a randomly initialized encoder with the same backbone. Tables~\ref{tab:pretraining}--\ref{tab:pretraining_gains} separate absolute scores from differences in depth and residual gains.}
\input{tables/extra_pretraining}
\input{tables/extra_pretraining_gains}

%% file: tables/supp_table_style.tex
\newcommand{\supptable}{%
  \fontsize{9}{10.5}\selectfont
  \renewcommand{\arraystretch}{1.1}%
  \setlength{\tabcolsep}{4pt}%
  \setlength{\abovecaptionskip}{7pt}%
  \setlength{\belowcaptionskip}{7pt}%
}
\newcommand{\suppci}[3]{#1\,{\fontsize{8.5}{10}\selectfont[#2, #3]}}
\newcommand{\suppcistack}[3]{\raisebox{-0.35\height}{\shortstack[c]{#1\\[1pt]{\fontsize{8}{9}\selectfont[#2, #3]}}}}
\newcommand{\supppanel}[1]{\par{\raggedright\textbf{#1}\par}\vspace{4pt}}

%% file: tables/b_absolute.tex
\begin{table}[H]
\centering\supptable\renewcommand{\arraystretch}{1.03}

\caption{\rev{Absolute scores (\%) and within-task ranks in parentheses. BA: OpenBMI; AUROC: N170/N400. Readout is selected at the final layer, then depth is selected at that readout.}}
\label{tab:depth}
\label{tab:rankings}
\begin{tabular*}{\linewidth}{@{\extracolsep{\fill}}llrrr@{}}
\toprule
Task & Model & Final mean & Final $r_L$ & \rev{Selected} / $r_L$ \\
\midrule
OpenBMI & CBraMod & 50.42 \rev{(5)} & 76.09 \rev{(2)} & 76.17 \rev{(2)} \\
 & REVE & 59.73 \rev{(1)} & 76.95 \rev{(1)} & 80.46 \rev{(1)} \\
 & LaBraM & 51.42 \rev{(4)} & 69.74 \rev{(4)} & 75.99 \rev{(3)} \\
 & LUNA & 55.37 \rev{(2)} & 59.67 \rev{(5)} & 59.71 \rev{(5)} \\
 & CSBrain & 52.65 \rev{(3)} & 71.20 \rev{(3)} & 75.10 \rev{(4)} \\
\addlinespace[3pt]
N170 & CBraMod & 57.10 \rev{(3)} & 68.77 \rev{(3)} & 77.45 \rev{(3)} \\
 & REVE & 70.91 \rev{(2)} & 73.41 \rev{(2)} & 80.81 \rev{(1)} \\
 & LaBraM & 56.07 \rev{(4)} & 67.01 \rev{(4)} & 75.12 \rev{(4)} \\
 & LUNA & 52.40 \rev{(5)} & 58.25 \rev{(5)} & 62.79 \rev{(5)} \\
 & CSBrain & 71.42 \rev{(1)} & 74.78 \rev{(1)} & 79.09 \rev{(2)} \\
\addlinespace[3pt]
N400 & CBraMod & 62.61 \rev{(4)} & 71.18 \rev{(4)} & 77.59 \rev{(3)} \\
 & REVE & 72.58 \rev{(2)} & 75.81 \rev{(1)} & 78.48 \rev{(1)} \\
 & LaBraM & 65.95 \rev{(3)} & 70.87 \rev{(5)} & 76.04 \rev{(4)} \\
 & LUNA & 58.70 \rev{(5)} & 71.57 \rev{(3)} & 71.85 \rev{(5)} \\
 & CSBrain & 73.26 \rev{(1)} & 75.52 \rev{(2)} & 77.99 \rev{(2)} \\
\bottomrule
\end{tabular*}
\end{table}

%% file: tables/b_full.tex
\begin{table}[H]
\centering\supptable\renewcommand{\arraystretch}{1.03}

\caption{Complete primary effects (percent, paired 95\% intervals). Readout correction, fixed-readout depth utility, and joint interface utility remain separate quantities.}
\label{tab:full}
\begin{tabular*}{\linewidth}{@{\extracolsep{\fill}}llrrr@{}}
\toprule
Task & Model & $\Delta_{\rm readout}$ & $\udepth$ & $\uinterface$ \\
\midrule
OpenBMI & CBraMod & \suppci{25.67}{23.19}{28.17} & \suppci{0.07}{-0.61}{0.74} & \suppci{0.07}{-0.61}{0.74} \\
 & REVE & \suppci{17.23}{15.54}{18.99} & \suppci{3.50}{2.32}{4.72} & \suppci{3.50}{2.32}{4.72} \\
 & LaBraM & \suppci{18.32}{16.07}{20.52} & \suppci{6.25}{5.46}{7.06} & \suppci{6.25}{5.46}{7.06} \\
 & LUNA & \suppci{4.30}{2.72}{5.93} & \suppci{0.04}{-0.30}{0.39} & \suppci{0.04}{-0.30}{0.39} \\
 & CSBrain & \suppci{18.55}{16.40}{20.80} & \suppci{3.91}{3.11}{4.70} & \suppci{3.91}{3.11}{4.70} \\
\addlinespace[3pt]
N170 & CBraMod & \suppci{11.68}{9.28}{14.07} & \suppci{8.67}{6.86}{10.45} & \suppci{8.67}{6.86}{10.45} \\
 & REVE & \suppci{2.50}{1.10}{3.95} & \suppci{7.40}{5.90}{8.94} & \suppci{7.40}{5.90}{8.94} \\
 & LaBraM & \suppci{10.94}{8.35}{13.48} & \suppci{8.11}{5.69}{10.78} & \suppci{8.11}{5.69}{10.78} \\
 & LUNA & \suppci{5.85}{4.00}{7.82} & \suppci{4.53}{2.54}{6.71} & \suppci{4.53}{2.54}{6.71} \\
 & CSBrain & \suppci{3.35}{1.61}{5.11} & \suppci{4.32}{3.24}{5.45} & \suppci{4.32}{3.24}{5.45} \\
\addlinespace[3pt]
N400 & CBraMod & \suppci{8.57}{6.69}{10.39} & \suppci{6.41}{5.23}{7.61} & \suppci{6.41}{5.23}{7.61} \\
 & REVE & \suppci{3.23}{2.10}{4.36} & \suppci{2.67}{1.59}{3.76} & \suppci{2.67}{1.59}{3.76} \\
 & LaBraM & \suppci{4.93}{3.52}{6.34} & \suppci{5.16}{4.00}{6.37} & \suppci{5.16}{4.00}{6.37} \\
 & LUNA & \suppci{12.87}{10.87}{14.93} & \suppci{0.28}{-0.08}{0.66} & \suppci{0.28}{-0.08}{0.66} \\
 & CSBrain & \suppci{2.26}{1.23}{3.30} & \suppci{2.47}{1.69}{3.21} & \suppci{2.47}{1.69}{3.21} \\
\addlinespace[3pt]
SEED-V & CBraMod & \suppci{5.00}{1.11}{8.75} & \suppci{0.28}{-2.78}{3.61} & \suppci{1.53}{-1.11}{4.44} \\
 & REVE & \suppci{-1.67}{-5.69}{2.08} & \suppci{0.69}{-1.94}{3.19} & \suppci{1.53}{-3.61}{6.81} \\
 & LaBraM & \suppci{3.47}{0.28}{6.67} & \suppci{-2.50}{-5.97}{0.83} & \suppci{-2.36}{-6.11}{1.53} \\
 & LUNA & \suppci{1.25}{-1.11}{3.61} & \suppci{-0.56}{-4.31}{2.92} & \suppci{-0.28}{-4.31}{3.47} \\
 & CSBrain & \suppci{0.28}{-2.22}{3.06} & \suppci{1.11}{-1.53}{4.03} & \suppci{-0.69}{-2.50}{1.11} \\
\bottomrule
\end{tabular*}
\end{table}

%% file: tables/b_learned.tex
\begin{table}[H]
\centering\supptable

\caption{Learned-family controls. Absolute scores are percentages; selected-minus-final gains are expressed in percent with paired 95\% intervals. Seeds are averaged within participant before bootstrap.}
\label{tab:learned}
\begin{tabular*}{\linewidth}{@{\extracolsep{\fill}}lllrrr@{}}
\toprule
Family & Model & Task & Final & Selected & Gain [95\% interval] \\
\midrule
Global query & CBraMod & N170 & 62.96 & 72.16 & \suppci{9.20}{6.14}{12.36} \\
 & CBraMod & OpenBMI & 56.61 & 60.59 & \suppci{3.98}{2.55}{5.49} \\
 & REVE & N170 & 69.74 & 73.92 & \suppci{4.18}{1.86}{6.22} \\
 & REVE & OpenBMI & 65.91 & 70.88 & \suppci{4.97}{3.69}{6.40} \\
\addlinespace[4pt]
Channel-preserving & CBraMod & N170 & 68.38 & 77.15 & \suppci{8.77}{7.05}{10.48} \\
 & CBraMod & OpenBMI & 71.32 & 74.90 & \suppci{3.57}{2.58}{4.52} \\
 & REVE & N170 & 72.64 & 80.50 & \suppci{7.86}{6.22}{9.56} \\
 & REVE & OpenBMI & 77.05 & 79.86 & \suppci{2.81}{1.48}{4.17} \\
\bottomrule
\end{tabular*}
\end{table}

%% file: tables/b_budget.tex
\begin{table}[H]
\centering\supptable

\caption{Full minus 10\% fitting-label effects (percent, paired intervals). Validation pools stay fixed. The second quantity is joint interface utility.}
\label{tab:budget}
\begin{tabular*}{\linewidth}{@{\extracolsep{\fill}}llrr@{}}
\toprule
Task & Model & Change in $\Delta_{\rm readout}$ & Change in $\uinterface$ \\
\midrule
OpenBMI & CBraMod & \suppci{8.82}{7.46}{10.14} & \suppci{0.38}{-0.30}{1.08} \\
 & LaBraM & \suppci{8.77}{7.49}{10.06} & \suppci{1.27}{0.30}{2.27} \\
 & CSBrain & \suppci{8.43}{7.00}{9.93} & \suppci{1.91}{0.85}{2.95} \\
\addlinespace[4pt]
N170 & CBraMod & \suppci{7.43}{5.36}{9.63} & \suppci{-1.38}{-3.31}{0.51} \\
 & LaBraM & \suppci{4.24}{2.32}{6.23} & \suppci{2.96}{1.41}{4.71} \\
 & CSBrain & \suppci{1.54}{-0.47}{3.77} & \suppci{0.62}{-0.92}{2.13} \\
\bottomrule
\end{tabular*}
\end{table}

%% file: tables/b_stability.tex
\begin{table}[H]
\centering\supptable
\caption{\rev{Selection stability across five outer folds. Each cell lists the jointly selected layer indices; parentheses give the mean development-score margin over the runner-up (\%).}}
\label{tab:stability}
\begin{tabular*}{\linewidth}{@{\extracolsep{\fill}}lrrrr@{}}\toprule
Model & OpenBMI & N170 & N400 & SEED-V \\
\midrule
CBraMod & \shortstack{8, 12, 9, 7, 4\\(0.169)} & \shortstack{8, 2, 4, 2, 4\\(0.227)} & \shortstack{1, 5, 2, 3, 4\\(0.184)} & \shortstack{9, 10, 1, 8, 9\\(0.350)} \\
REVE & \shortstack{11, 11, 11, 10, 11\\(0.194)} & \shortstack{7, 7, 8, 8, 8\\(0.078)} & \shortstack{10, 10, 9, 9, 9\\(0.168)} & \shortstack{11, 20, 14, 20, 10\\(0.698)} \\
LaBraM & \shortstack{7, 4, 6, 4, 7\\(0.226)} & \shortstack{1, 1, 2, 1, 2\\(0.251)} & \shortstack{4, 4, 7, 4, 4\\(0.109)} & \shortstack{12, 1, 9, 5, 5\\(0.450)} \\
LUNA & \shortstack{8, 8, 8, 7, 8\\(0.213)} & \shortstack{1, 3, 1, 1, 1\\(0.368)} & \shortstack{7, 8, 7, 7, 7\\(0.338)} & \shortstack{5, 3, 2, 1, 2\\(0.490)} \\
CSBrain & \shortstack{3, 3, 3, 3, 3\\(0.528)} & \shortstack{1, 1, 2, 2, 2\\(0.219)} & \shortstack{3, 4, 5, 4, 4\\(0.198)} & \shortstack{8, 4, 3, 1, 10\\(0.427)} \\
\bottomrule\end{tabular*}
\end{table}

%% file: tables/f_crossfit.tex
\begin{table}[H]
\centering\supptable\renewcommand{\arraystretch}{1.0}
\caption{Cross-fitted forward utility and shuffled residuals of the same dimension (percent; paired 95\% intervals). Permutation scores are averaged over 20 seeds within participant before resampling. Both $\Delta_{\rm resid}$ and $E_{\rm resid}$ have positive intervals in 13/15 conditions.}
\label{tab:crossfit}
\begin{tabular*}{\linewidth}{@{\extracolsep{\fill}}llrrr@{}}
\toprule
Task & Model & $\Delta_{\rm resid}$ & $\overline\Delta_{\rm perm}$ & $E_{\rm resid}$ \\
\midrule
OpenBMI & CBraMod & \suppci{0.84}{0.19}{1.53} & \suppci{-0.81}{-1.13}{-0.50} & \suppci{1.65}{1.06}{2.28} \\
 & REVE & \suppci{7.25}{6.07}{8.45} & \suppci{-0.23}{-0.54}{0.08} & \suppci{7.48}{6.36}{8.60} \\
 & LaBraM & \suppci{5.00}{4.03}{5.99} & \suppci{-0.51}{-0.86}{-0.16} & \suppci{5.50}{4.54}{6.45} \\
 & LUNA & \suppci{-0.01}{-0.53}{0.52} & \suppci{-0.23}{-0.46}{-0.00} & \suppci{0.22}{-0.28}{0.74} \\
 & CSBrain & \suppci{3.74}{2.71}{4.80} & \suppci{-0.41}{-0.83}{0.02} & \suppci{4.15}{3.13}{5.19} \\
\addlinespace[4pt]
N170 & CBraMod & \suppci{10.03}{7.92}{12.17} & \suppci{-1.74}{-2.16}{-1.32} & \suppci{11.77}{9.65}{13.94} \\
 & REVE & \suppci{9.24}{7.34}{11.11} & \suppci{-1.75}{-2.06}{-1.46} & \suppci{10.99}{9.13}{12.82} \\
 & LaBraM & \suppci{6.78}{4.84}{8.70} & \suppci{-1.90}{-2.31}{-1.48} & \suppci{8.69}{6.68}{10.66} \\
 & LUNA & \suppci{3.44}{1.66}{5.49} & \suppci{-1.58}{-2.11}{-1.08} & \suppci{5.03}{3.38}{6.93} \\
 & CSBrain & \suppci{3.97}{2.86}{5.10} & \suppci{-1.33}{-1.62}{-1.04} & \suppci{5.30}{4.20}{6.43} \\
\addlinespace[4pt]
N400 & CBraMod & \suppci{6.18}{4.70}{7.68} & \suppci{-1.73}{-2.02}{-1.45} & \suppci{7.91}{6.41}{9.46} \\
 & REVE & \suppci{2.25}{1.13}{3.31} & \suppci{-1.66}{-1.95}{-1.37} & \suppci{3.91}{2.79}{5.02} \\
 & LaBraM & \suppci{5.26}{3.91}{6.58} & \suppci{-2.32}{-2.69}{-1.94} & \suppci{7.58}{6.27}{8.89} \\
 & LUNA & \suppci{-0.97}{-1.71}{-0.25} & \suppci{-1.52}{-1.91}{-1.15} & \suppci{0.55}{-0.14}{1.23} \\
 & CSBrain & \suppci{1.24}{0.07}{2.46} & \suppci{-1.74}{-1.98}{-1.50} & \suppci{2.99}{1.80}{4.22} \\
\bottomrule
\end{tabular*}
\end{table}

%% file: tables/f_crossfit_checks.tex
\begin{table}[H]
\centering\supptable\renewcommand{\arraystretch}{1.0}

\caption{\rev{Cross-fitted representation scores (\%) with marginal 95\% participant-bootstrap intervals. Final and Selected each use 200 PCA features; Final + residual uses 400.}}
\label{tab:crossfitchecks}
\begin{tabular*}{\linewidth}{@{\extracolsep{\fill}}llrrrr@{}}
\toprule
Task & Model & \rev{Final} & \rev{Selected} & \rev{Residual} & \rev{Final + residual} \\
\midrule
OpenBMI & CBraMod & \suppci{70.28}{67.89}{72.68} & \suppci{68.30}{65.94}{70.81} & \suppci{54.27}{52.98}{55.62} & \suppci{71.12}{68.84}{73.50} \\
 & REVE & \suppci{66.39}{64.12}{68.81} & \suppci{74.80}{72.30}{77.32} & \suppci{63.66}{61.53}{65.91} & \suppci{73.65}{71.13}{76.22} \\
 & LaBraM & \suppci{61.87}{59.64}{64.25} & \suppci{66.45}{64.01}{68.96} & \suppci{59.44}{57.67}{61.36} & \suppci{66.87}{64.57}{69.29} \\
 & LUNA & \suppci{59.06}{57.08}{61.11} & \suppci{59.09}{57.17}{61.13} & \suppci{52.07}{51.01}{53.23} & \suppci{59.05}{57.04}{61.11} \\
 & CSBrain & \suppci{60.70}{58.35}{63.25} & \suppci{63.56}{61.25}{66.00} & \suppci{58.26}{57.00}{59.54} & \suppci{64.44}{61.93}{67.07} \\
\addlinespace[4pt]
N170 & CBraMod & \suppci{64.71}{62.14}{67.32} & \suppci{74.74}{71.36}{78.12} & \suppci{68.56}{65.07}{72.18} & \suppci{74.74}{71.55}{77.93} \\
 & REVE & \suppci{68.12}{65.42}{70.83} & \suppci{78.40}{74.98}{81.75} & \suppci{69.16}{66.17}{72.22} & \suppci{77.35}{74.11}{80.55} \\
 & LaBraM & \suppci{64.72}{62.11}{67.31} & \suppci{72.29}{68.80}{75.73} & \suppci{64.69}{61.38}{68.04} & \suppci{71.50}{68.28}{74.76} \\
 & LUNA & \suppci{57.37}{55.68}{59.06} & \suppci{61.48}{59.56}{63.50} & \suppci{57.71}{55.74}{59.85} & \suppci{60.81}{58.95}{62.71} \\
 & CSBrain & \suppci{71.47}{67.94}{74.99} & \suppci{75.69}{72.03}{79.33} & \suppci{61.40}{59.18}{63.62} & \suppci{75.44}{72.07}{78.73} \\
\addlinespace[4pt]
N400 & CBraMod & \suppci{69.33}{67.08}{71.64} & \suppci{76.79}{74.20}{79.39} & \suppci{66.14}{63.75}{68.62} & \suppci{75.51}{72.93}{78.16} \\
 & REVE & \suppci{73.78}{71.16}{76.44} & \suppci{77.47}{75.10}{79.87} & \suppci{60.46}{58.34}{62.63} & \suppci{76.03}{73.33}{78.74} \\
 & LaBraM & \suppci{68.61}{66.33}{70.93} & \suppci{75.25}{72.94}{77.54} & \suppci{64.66}{62.56}{66.94} & \suppci{73.87}{71.45}{76.29} \\
 & LUNA & \suppci{70.03}{67.55}{72.55} & \suppci{70.17}{67.62}{72.69} & \suppci{51.90}{50.04}{53.66} & \suppci{69.06}{66.69}{71.44} \\
 & CSBrain & \suppci{74.31}{72.06}{76.57} & \suppci{77.25}{74.91}{79.60} & \suppci{58.69}{56.64}{60.72} & \suppci{75.55}{73.26}{77.91} \\
\bottomrule
\end{tabular*}
\end{table}

%% file: tables/f_sensitivity.tex
\begin{table}[H]
\centering\supptable

\caption{\rev{Original sensitivity results: augmentation minus Final (\%) [95\% CI]. Uncompressed ERP features retain all four negative results; Gaussian projections use 200 features per layer. These runs are distinct from the cross-fitted dimension scan.}}
\label{tab:v5sensitivity}
\begin{tabular*}{\linewidth}{@{\extracolsep{\fill}}lllr@{}}
\toprule
Task & Model & \rev{Representation} & \rev{Augmentation gain} \\
\midrule
N170 & CBraMod & \rev{Uncompressed} & \suppci{-6.18}{-9.54}{-3.15} \\
N170 & REVE & \rev{Uncompressed} & \suppci{-14.35}{-17.57}{-11.15} \\
N400 & CBraMod & \rev{Uncompressed} & \suppci{-9.81}{-12.04}{-7.59} \\
N400 & REVE & \rev{Uncompressed} & \suppci{-18.32}{-21.32}{-15.41} \\
N170 & CBraMod & \rev{Gaussian projection (200)} & \suppci{9.63}{7.19}{12.16} \\
N170 & REVE & \rev{Gaussian projection (200)} & \suppci{7.52}{5.37}{9.54} \\
N400 & CBraMod & \rev{Gaussian projection (200)} & \suppci{6.83}{5.18}{8.44} \\
N400 & REVE & \rev{Gaussian projection (200)} & \suppci{2.88}{1.65}{4.16} \\
OpenBMI & CBraMod & \rev{Gaussian projection (200)} & \suppci{2.53}{1.50}{3.55} \\
OpenBMI & REVE & \rev{Gaussian projection (200)} & \suppci{3.79}{2.72}{4.82} \\
\bottomrule
\end{tabular*}
\end{table}

%% file: tables/extra_dimensions.tex
\begin{table}[H]
\centering\color{RevisionBlue}\supptable
\caption{\rev{Cross-fitted dimension scan: AUROC gains (\%) [paired 95\% CI]. All conditions use ridge mapping and a linear task head. Full denotes 5,600 features for CBraMod and 14,336 for REVE.}}
\label{tab:dimension}
\textbf{A. Augmentation minus Final}\par\smallskip
\begin{tabular*}{\linewidth}{@{\extracolsep{\fill}}rrrrrr@{}}\toprule
\shortstack{Final\\features} & \shortstack{Selected\\features} & \shortstack{CBraMod\\N170} & \shortstack{REVE\\N170} & \shortstack{CBraMod\\N400} & \shortstack{REVE\\N400} \\
\midrule
100 & 100 & \suppci{11.05}{7.74}{13.91} & \suppci{8.83}{6.93}{10.62} & \suppci{8.60}{6.87}{10.33} & \suppci{2.55}{1.48}{3.66} \\
200 & 200 & \suppci{9.57}{7.33}{11.83} & \suppci{8.96}{7.04}{10.87} & \suppci{6.11}{4.55}{7.66} & \suppci{2.12}{1.14}{3.07} \\
400 & 400 & \suppci{7.62}{5.00}{9.89} & \suppci{7.59}{6.03}{9.16} & \suppci{4.59}{3.06}{6.14} & \suppci{-0.36}{-1.25}{0.51} \\
400 & 200 & \suppci{6.87}{4.81}{8.88} & \suppci{6.28}{4.96}{7.65} & \suppci{5.24}{3.81}{6.72} & \suppci{0.53}{-0.18}{1.29} \\
Full & 200 & \suppci{1.22}{0.37}{2.02} & \suppci{0.48}{0.31}{0.66} & \suppci{0.71}{0.44}{0.98} & \suppci{-0.01}{-0.15}{0.11} \\
Full & Full & \suppci{-0.16}{-2.43}{1.88} & \suppci{0.25}{-0.81}{1.37} & \suppci{0.14}{-0.88}{1.12} & \suppci{-0.99}{-1.94}{-0.08} \\
\bottomrule\end{tabular*}
\par\medskip
\textbf{B. Augmentation minus shuffled augmentation}\par\smallskip
\begin{tabular*}{\linewidth}{@{\extracolsep{\fill}}rrrrrr@{}}\toprule
\shortstack{Final\\features} & \shortstack{Selected\\features} & \shortstack{CBraMod\\N170} & \shortstack{REVE\\N170} & \shortstack{CBraMod\\N400} & \shortstack{REVE\\N400} \\
\midrule
100 & 100 & \suppci{12.08}{9.51}{14.61} & \suppci{10.75}{8.88}{12.47} & \suppci{10.06}{8.08}{12.07} & \suppci{4.13}{2.96}{5.28} \\
200 & 200 & \suppci{12.07}{10.29}{13.94} & \suppci{11.13}{8.92}{13.32} & \suppci{8.15}{6.36}{9.94} & \suppci{3.87}{2.70}{5.05} \\
400 & 400 & \suppci{10.28}{8.17}{12.21} & \suppci{9.89}{8.49}{11.35} & \suppci{6.43}{4.47}{8.39} & \suppci{1.69}{0.42}{2.94} \\
400 & 200 & \suppci{8.94}{7.41}{10.46} & \suppci{7.51}{6.15}{8.85} & \suppci{6.24}{4.62}{7.90} & \suppci{1.87}{0.90}{2.94} \\
Full & 200 & \suppci{1.88}{1.00}{2.72} & \suppci{0.57}{0.36}{0.81} & \suppci{0.72}{0.39}{1.07} & \suppci{-0.01}{-0.17}{0.13} \\
Full & Full & \suppci{3.22}{0.47}{5.49} & \suppci{3.05}{1.81}{4.34} & \suppci{1.76}{0.25}{3.25} & \suppci{0.77}{-0.31}{1.88} \\
\bottomrule\end{tabular*}
\end{table}

%% file: tables/extra_fusion.tex
\begin{table}[H]
\centering\color{RevisionBlue}\supptable

\caption{\rev{Decoder comparison. Panel A fixes the total classifier input at 200 features; bold marks the largest score in each row. Panel B uses unequal feature counts. Three-layer methods use the first, middle, and final layers.}}
\label{tab:fusion}
\textbf{A. Matched total feature budget}\par\smallskip
\begin{tabular*}{\linewidth}{@{\extracolsep{\fill}}llrrrrrr@{}}
\toprule
Task & Model & Final & Selected & \shortstack{Final +\\residual} & \shortstack{Final +\\selected} & \shortstack{Three-layer\\concat.} & \shortstack{Learned\\weights} \\
\midrule
OpenBMI & CBraMod & \textbf{69.86} & 68.22 & 68.55 & 68.41 & 65.76 & 66.15 \\
OpenBMI & REVE & 66.37 & \textbf{74.14} & 70.11 & 70.68 & 67.28 & 69.17 \\
N170 & CBraMod & 64.85 & \textbf{74.59} & 70.77 & 71.37 & 71.23 & 74.28 \\
N170 & REVE & 68.23 & \textbf{78.24} & 74.37 & 75.03 & 75.03 & 71.92 \\
\bottomrule
\end{tabular*}
\par\medskip\textbf{B. Uncompressed concatenation}\par\smallskip
\begin{tabular*}{\linewidth}{@{\extracolsep{\fill}}llrr@{}}\toprule
Task & Model & Selected (200) & Three-layer concatenation (full) \\\midrule
OpenBMI & CBraMod & 68.22 & 76.19 \\
OpenBMI & REVE & 74.14 & 79.05 \\
N170 & CBraMod & 74.59 & 77.70 \\
N170 & REVE & 78.24 & 78.87 \\
\bottomrule\end{tabular*}\par
\end{table}

%% file: tables/supp_capacity.tex
\begin{table}[H]
\centering\color{RevisionBlue}\supptable

\caption{\rev{Mapping and task-head capacity at 200 features per layer. Scores and gains are percentages; brackets give paired 95\% intervals. BA: OpenBMI; AUROC: N170/N400. The same head family is used within each comparison.}}
\label{tab:capacity}
\begin{tabular*}{\linewidth}{@{\extracolsep{\fill}}llrrrr@{}}
\toprule
Mapper & Head & Final & Selected & \shortstack{Augmentation\\minus Final} & \shortstack{Augmentation\\minus shuffled} \\
\midrule
\multicolumn{6}{l}{\textbf{OpenBMI / CBraMod}} \\
Ridge & Linear & 69.86 & 68.40 & \suppci{1.52}{0.92}{2.10} & \suppci{2.04}{1.21}{2.83} \\
Ridge & MLP & 69.76 & 68.33 & \suppci{1.36}{0.81}{1.93} & \suppci{1.87}{1.23}{2.51} \\
MLP (256) & Linear & 69.86 & 68.40 & \suppci{0.71}{0.08}{1.32} & \suppci{0.95}{0.18}{1.70} \\
MLP (256) & MLP & 69.76 & 68.33 & \suppci{0.84}{0.27}{1.39} & \suppci{0.88}{0.24}{1.52} \\
MLP (512, 512) & Linear & 69.86 & 68.40 & \suppci{0.71}{0.07}{1.34} & \suppci{1.16}{0.34}{1.96} \\
MLP (512, 512) & MLP & 69.76 & 68.33 & \suppci{0.94}{0.33}{1.56} & \suppci{1.33}{0.60}{2.08} \\
\addlinespace[3pt]
\multicolumn{6}{l}{\textbf{N170 / CBraMod}} \\
Ridge & Linear & 64.85 & 74.46 & \suppci{9.57}{7.33}{11.83} & \suppci{12.07}{10.29}{13.94} \\
Ridge & MLP & 64.33 & 74.16 & \suppci{9.91}{7.73}{12.06} & \suppci{12.26}{10.28}{14.29} \\
MLP (256) & Linear & 64.85 & 74.46 & \suppci{8.39}{6.01}{10.74} & \suppci{10.33}{8.31}{12.50} \\
MLP (256) & MLP & 64.33 & 74.16 & \suppci{7.81}{5.47}{10.05} & \suppci{10.11}{8.25}{12.08} \\
MLP (512, 512) & Linear & 64.85 & 74.46 & \suppci{9.87}{7.81}{11.96} & \suppci{12.34}{10.37}{14.40} \\
MLP (512, 512) & MLP & 64.33 & 74.16 & \suppci{9.83}{7.71}{11.94} & \suppci{12.06}{9.97}{14.22} \\
\addlinespace[3pt]
\multicolumn{6}{l}{\textbf{N400 / CSBrain}} \\
Ridge & Linear & 74.24 & 77.22 & \suppci{1.34}{0.34}{2.32} & \suppci{2.81}{1.48}{4.10} \\
Ridge & MLP & 74.11 & 77.15 & \suppci{1.38}{0.34}{2.41} & \suppci{3.16}{1.79}{4.50} \\
MLP (256) & Linear & 74.24 & 77.22 & \suppci{0.17}{-0.70}{1.01} & \suppci{1.71}{0.52}{2.86} \\
MLP (256) & MLP & 74.11 & 77.15 & \suppci{0.01}{-0.87}{0.88} & \suppci{1.78}{0.47}{3.02} \\
MLP (512, 512) & Linear & 74.24 & 77.22 & \suppci{1.50}{0.48}{2.47} & \suppci{3.29}{1.81}{4.61} \\
MLP (512, 512) & MLP & 74.11 & 77.15 & \suppci{1.19}{0.23}{2.10} & \suppci{2.99}{1.50}{4.38} \\
\addlinespace[3pt]
\bottomrule
\end{tabular*}
\end{table}

%% file: tables/extra_mapper.tex
\begin{table}[H]
\centering\color{RevisionBlue}\supptable

\caption{\rev{Held-out reconstruction diagnostics at PCA200: unweighted means over the five outer-fold mappings. These are reconstruction $R^2$/MSE, not downstream task scores.}}
\label{tab:mapper}
\begin{tabular*}{\linewidth}{@{\extracolsep{\fill}}lllrr@{}}
\toprule
Task & Model & Mapper & $R^2$ & MSE \\
\midrule
OpenBMI & CBraMod & Ridge & 0.804 & 20.069 \\
 & CBraMod & MLP 256 & 0.757 & 24.035 \\
 & CBraMod & MLP 512/512 & 0.706 & 29.197 \\
\addlinespace[4pt]
N170 & CBraMod & Ridge & 0.566 & 7.051 \\
 & CBraMod & MLP 256 & 0.478 & 8.453 \\
 & CBraMod & MLP 512/512 & 0.464 & 8.671 \\
\addlinespace[4pt]
N400 & CSBrain & Ridge & 0.721 & 5.899 \\
 & CSBrain & MLP 256 & 0.615 & 8.150 \\
 & CSBrain & MLP 512/512 & 0.600 & 8.461 \\
\bottomrule
\end{tabular*}
\end{table}

%% file: tables/extra_selection.tex
\begin{table}[H]

\centering\color{RevisionBlue}\supptable

\caption{\rev{Fixed-layer controls under the same source readout (\%). Sleep uses macro-F1 and MI uses BA. First and Middle supplement the Final, Random, Source, and Target strategies in Table~\ref{tab:transfer}.}}
\label{tab:selection_extra}
\begin{tabular*}{\linewidth}{@{\extracolsep{\fill}}llrrrrrr@{}}
\toprule
Task & Model & Final & First & Middle & Random & Source & Target \\
\midrule
Sleep & CBraMod & 70.56 & 70.97 & 71.76 & 71.12 & 71.76 & 71.56 \\
 & REVE & 71.66 & 59.83 & 72.18 & 69.30 & 72.79 & 72.97 \\
 & LaBraM & 69.22 & 67.14 & 70.22 & 69.65 & 70.29 & 70.27 \\
\addlinespace[4pt]
MI full & CBraMod & 66.98 & 71.99 & 69.06 & 69.43 & 71.99 & 71.72 \\
 & REVE & 77.08 & 71.22 & 79.82 & 76.10 & 79.82 & 79.82 \\
\bottomrule
\end{tabular*}
\end{table}

%% file: tables/transfer_gains.tex
\begin{table}[H]

\centering\color{RevisionBlue}\supptable

\caption{\rev{Source-selected gains across five dataset pairs (\%). $\Delta_F$: Source minus Final; $\Delta_R$: Source minus Random. Sleep uses macro-F1; other tasks use BA. $^*$Paired 95\% CI includes zero. Full intervals and pair-specific readouts appear in Appendix~\ref{app:transfer_extra}.}}
\label{tab:transfer_gains}
\begin{tabular*}{\linewidth}{@{\extracolsep{\fill}}lrrrrr@{}}
\toprule
Transfer / task & $n$ & \shortstack{CBraMod\\$\Delta_F$} & \shortstack{CBraMod\\$\Delta_R$} & \shortstack{REVE\\$\Delta_F$} & \shortstack{REVE\\$\Delta_R$} \\
\midrule
HMC--ISRUC / Sleep & 98 & 1.20 & 0.64 & 1.13 & 3.48 \\
OpenBMI--BCIC-IV-2a / MI & 9 & 5.02 & 2.56 & 2.74 & 3.72 \\
OpenBMI--WBCIC-SHU / MI & 51 & 1.45 & 0.31$^*$ & 0.88 & 0.77 \\
Benchmark--BETA / SSVEP & 69 & 9.39 & 3.54 & 12.75 & 3.65 \\
ERP CORE P3--Won2022 / P300 & 55 & 4.72 & 0.67 & 1.11 & 0.07$^*$ \\
\bottomrule
\end{tabular*}
\end{table}

%% file: tables/extra_transfer_ci.tex
\begin{table}[H]

\centering\color{RevisionBlue}\supptable

\caption{\rev{Paired source-selected transfer gains (\%) [95\% CI] for the five pairs in Table~\ref{tab:transfer_gains}.}}
\label{tab:transfer_ci}
\begin{tabular*}{\linewidth}{@{\extracolsep{\fill}}llrrr@{}}
\toprule
Target & Model & \shortstack{Source minus\\Final} & \shortstack{Source minus\\Random} & \shortstack{Source minus\\First} \\
\midrule
Sleep & CBraMod & \suppci{1.20}{0.63}{1.78} & \suppci{0.64}{0.28}{1.01} & \suppci{0.79}{0.06}{1.64} \\
 & REVE & \suppci{1.13}{0.49}{1.84} & \suppci{3.48}{2.72}{4.32} & \suppci{12.96}{10.99}{15.06} \\
\addlinespace[4pt]
BCIC-IV-2a & CBraMod & \suppci{5.02}{3.40}{6.37} & \suppci{2.56}{1.80}{3.21} & \suppci{0.00}{0.00}{0.00} \\
 & REVE & \suppci{2.74}{0.50}{5.25} & \suppci{3.72}{2.39}{4.98} & \suppci{8.60}{3.63}{13.27} \\
\addlinespace[4pt]
WBCIC-SHU & CBraMod & \suppci{1.45}{0.87}{2.03} & \suppci{0.31}{-0.02}{0.64} & \suppci{0.04}{-0.42}{0.51} \\
 & REVE & \suppci{0.88}{0.33}{1.42} & \suppci{0.77}{0.40}{1.13} & \suppci{1.51}{0.65}{2.44} \\
\addlinespace[4pt]
BETA & CBraMod & \suppci{9.39}{7.55}{11.32} & \suppci{3.54}{2.87}{4.25} & \suppci{0.00}{0.00}{0.00} \\
 & REVE & \suppci{12.75}{11.02}{14.57} & \suppci{3.65}{2.90}{4.43} & \suppci{1.59}{0.50}{2.71} \\
\addlinespace[4pt]
Won2022 & CBraMod & \suppci{4.72}{4.32}{5.12} & \suppci{0.67}{0.57}{0.77} & \suppci{0.24}{0.07}{0.42} \\
 & REVE & \suppci{1.11}{0.92}{1.30} & \suppci{0.07}{-0.04}{0.17} & \suppci{0.47}{0.21}{0.71} \\
\bottomrule
\end{tabular*}
\end{table}

%% file: tables/d_mi_followup.tex
\begin{table}[H]
\centering\supptable

\caption{\rev{Participant-level MI transfer in the original nine-participant evaluation. Entries are \rev{Source minus Final} BA (\%); summary intervals are paired 95\% bootstrap intervals. Source-selected layers are H01 for CBraMod and H11 for REVE.}}
\label{tab:mi_followup}
\label{tab:mi_reuse}
\begin{tabular*}{\linewidth}{@{\extracolsep{\fill}}lrr@{}}
\toprule
\rev{Participant / summary} & CBraMod & REVE \\
\midrule
A01 & +4.86 & -0.69 \\
A02 & +6.94 & +2.43 \\
A03 & +3.82 & +0.69 \\
A04 & +4.51 & +3.82 \\
A05 & +5.56 & +2.43 \\
A06 & +7.99 & +7.64 \\
A07 & +6.94 & -2.43 \\
A08 & -0.35 & +5.90 \\
A09 & +4.86 & +0.69 \\
\midrule
\rev{Final score} & \suppci{66.98}{61.73}{72.72} & \suppci{77.55}{72.80}{82.48} \\
\rev{Source score} & \suppci{71.99}{67.75}{76.66} & \suppci{79.82}{75.00}{84.72} \\
\rev{Source minus Final} & \suppci{5.02}{3.40}{6.40} & \suppci{2.28}{0.42}{4.24} \\
\rev{Positive participants} & 8/9 & 7/9 \\
\bottomrule
\end{tabular*}
\end{table}

%% file: tables/extra_crosstask.tex
\begin{table}[H]
\centering\color{RevisionBlue}\supptable

\caption{\rev{CBraMod cross-task depth reuse (\%) [95\% CI]. Target readout is selected on target development participants; only source layer identity transfers.}}
\label{tab:crosstask}
\begin{tabular*}{\linewidth}{@{\extracolsep{\fill}}lrrr@{}}
\toprule
Transfer & Source minus Final & Source minus Random & Source minus First \\
\midrule
OpenBMI$\to$N170 & \suppci{9.50}{7.59}{11.35} & \suppci{1.73}{0.87}{2.48} & \suppci{0.11}{-0.67}{0.87} \\
OpenBMI$\to$N400 & \suppci{6.76}{5.62}{7.89} & \suppci{1.89}{1.43}{2.32} & \suppci{0.12}{-0.75}{0.99} \\
N170$\to$OpenBMI & \suppci{-0.26}{-1.13}{0.60} & \suppci{0.00}{-0.44}{0.43} & \suppci{0.69}{0.02}{1.35} \\
\bottomrule
\end{tabular*}
\end{table}

%% file: tables/f_group2.tex
\begin{table}[H]
\centering\color{RevisionBlue}\supptable

\caption{\rev{HMC--ISRUC direction-transfer scores: macro-F1 (\%) [95\% CI], $n=98$. Source transformations are fixed. Deletion removes four directions at unit strength; random and shuffled controls average 20 independent bases.}}
\label{tab:v5group2}
\begin{tabular*}{\linewidth}{@{\extracolsep{\fill}}lrrr@{}}
\toprule
Representation / direction & CBraMod & REVE & LaBraM \\
\midrule
\textbf{Reference}: Final & \suppci{68.10}{65.85}{70.20} & \suppci{69.88}{67.57}{72.02} & \suppci{66.60}{64.31}{68.71} \\
\quad Selected & \suppci{69.09}{67.04}{71.06} & \suppci{71.45}{69.27}{73.50} & \suppci{68.22}{66.19}{70.04} \\
\addlinespace[2pt]
\textbf{Augmentation}: Task & \suppci{69.27}{67.04}{71.34} & \suppci{70.56}{68.27}{72.69} & \suppci{67.57}{65.35}{69.63} \\
\quad Random & \suppci{68.25}{66.02}{70.33} & \suppci{69.93}{67.64}{72.06} & \suppci{66.65}{64.38}{68.75} \\
\quad Residual PCA & \suppci{68.60}{66.44}{70.59} & \suppci{69.96}{67.67}{72.10} & \suppci{66.83}{64.56}{68.93} \\
\quad Shuffled & \suppci{68.37}{66.17}{70.43} & \suppci{69.93}{67.64}{72.06} & \suppci{66.63}{64.35}{68.74} \\
\quad Ordinary supervised & \suppci{68.72}{66.45}{70.83} & \suppci{70.12}{67.85}{72.20} & \suppci{66.99}{64.72}{69.10} \\
\quad Ordinary basis on residual & \suppci{68.76}{66.47}{70.87} & \suppci{70.01}{67.74}{72.09} & \suppci{66.99}{64.71}{69.10} \\
\addlinespace[2pt]
\textbf{Frozen deletion}: Task & \suppci{46.29}{44.13}{48.31} & \suppci{54.76}{52.49}{56.86} & \suppci{49.45}{46.95}{51.84} \\
\quad Random & \suppci{67.34}{65.01}{69.56} & \suppci{69.85}{67.54}{72.04} & \suppci{66.67}{64.48}{68.70} \\
\quad Residual PCA & \suppci{65.38}{62.87}{67.69} & \suppci{64.73}{61.49}{67.63} & \suppci{62.46}{59.63}{65.11} \\
\quad Shuffled & \suppci{66.10}{63.45}{68.62} & \suppci{68.90}{66.43}{71.23} & \suppci{65.76}{63.37}{68.00} \\
\addlinespace[2pt]
\textbf{Refit deletion}: Task & \suppci{67.30}{65.23}{69.31} & \suppci{70.63}{68.62}{72.51} & \suppci{65.95}{63.90}{67.79} \\
\quad Random & \suppci{69.00}{66.95}{70.97} & \suppci{71.41}{69.23}{73.48} & \suppci{68.12}{66.11}{69.93} \\
\quad Residual PCA & \suppci{68.34}{66.15}{70.43} & \suppci{70.90}{68.63}{73.03} & \suppci{67.79}{65.83}{69.58} \\
\quad Shuffled & \suppci{68.97}{66.92}{70.96} & \suppci{71.39}{69.21}{73.46} & \suppci{68.11}{66.08}{69.93} \\
\addlinespace[2pt]
\bottomrule
\end{tabular*}
\end{table}

%% file: tables/f_group2_controls.tex
\begin{table}[H]
\centering\color{RevisionBlue}\supptable

\caption{\rev{Paired direction-transfer contrasts (\%) [95\% CI]. Excess loss compares task-direction deletion with control deletion. The ordinary-basis comparison applies both direction bases to the same residual representation.}}
\label{tab:v5controls}
\begin{tabular*}{\linewidth}{@{\extracolsep{\fill}}lrrr@{}}
\toprule
Comparison & CBraMod & REVE & LaBraM \\
\midrule
Augmentation minus Final & \suppci{1.17}{0.72}{1.69} & \suppci{0.67}{0.30}{1.07} & \suppci{0.97}{0.67}{1.28} \\
Augmentation minus Random & \suppci{1.01}{0.60}{1.50} & \suppci{0.63}{0.27}{1.01} & \suppci{0.92}{0.63}{1.22} \\
Augmentation minus Shuffled & \suppci{0.89}{0.49}{1.34} & \suppci{0.62}{0.29}{0.98} & \suppci{0.94}{0.65}{1.24} \\
Augmentation minus Ordinary & \suppci{0.54}{0.13}{1.02} & \suppci{0.44}{0.05}{0.83} & \suppci{0.58}{0.26}{0.91} \\
Ordinary basis on same residual & \suppci{0.51}{0.08}{1.00} & \suppci{0.55}{0.13}{0.99} & \suppci{0.59}{0.26}{0.91} \\
Augmentation minus Selected & \suppci{0.18}{-0.68}{1.02} & \suppci{-0.89}{-1.49}{-0.30} & \suppci{-0.64}{-1.57}{0.17} \\
Refit excess loss / random & \suppci{1.70}{1.24}{2.20} & \suppci{0.78}{0.31}{1.28} & \suppci{2.17}{1.70}{2.65} \\
Refit excess loss / shuffled & \suppci{1.67}{1.21}{2.18} & \suppci{0.76}{0.29}{1.25} & \suppci{2.16}{1.69}{2.64} \\
Frozen excess loss / random & \suppci{21.04}{18.81}{23.33} & \suppci{15.08}{13.30}{16.99} & \suppci{17.22}{15.41}{19.11} \\
Frozen excess loss / shuffled & \suppci{19.81}{17.35}{22.28} & \suppci{14.14}{12.31}{16.06} & \suppci{16.31}{14.56}{18.14} \\
\bottomrule
\end{tabular*}
\end{table}

%% file: tables/extra_energy.tex
\begin{table}[H]
\centering\color{RevisionBlue}\supptable

\caption{\rev{Energy-matched perturbation scores (macro-F1, \%) [95\% CI], separate from equal-rank deletion. Scaling matches displacement energy after the target scaler and need not remove exactly four dimensions.}}
\label{tab:energy_extra}
\begin{tabular*}{\linewidth}{@{\extracolsep{\fill}}lrrr@{}}
\toprule
Perturbation & CBraMod & REVE & LaBraM \\
\midrule
Frozen / task & \suppci{46.29}{44.13}{48.31} & \suppci{54.76}{52.49}{56.86} & \suppci{49.45}{46.95}{51.84} \\
Frozen / random mean & \suppci{64.98}{62.38}{67.42} & \suppci{66.80}{64.20}{69.23} & \suppci{64.83}{62.39}{67.10} \\
Frozen / shuffle mean & \suppci{64.65}{61.83}{67.33} & \suppci{67.23}{64.56}{69.70} & \suppci{64.58}{61.99}{66.99} \\
Frozen / source pca & \suppci{62.87}{60.23}{65.25} & \suppci{60.37}{56.96}{63.45} & \suppci{58.90}{55.88}{61.79} \\
Refit / task & \suppci{67.30}{65.23}{69.31} & \suppci{70.63}{68.62}{72.51} & \suppci{65.95}{63.90}{67.79} \\
Refit / random mean & \suppci{69.09}{67.04}{71.06} & \suppci{71.46}{69.27}{73.52} & \suppci{68.23}{66.20}{70.04} \\
Refit / shuffle mean & \suppci{69.07}{67.02}{71.04} & \suppci{71.47}{69.28}{73.53} & \suppci{68.23}{66.20}{70.04} \\
Refit / source pca & \suppci{69.07}{67.02}{71.06} & \suppci{71.50}{69.32}{73.57} & \suppci{68.23}{66.21}{70.05} \\
\bottomrule
\end{tabular*}
\end{table}

%% file: tables/extra_physiology.tex
\begin{table}[H]
\centering\color{RevisionBlue}\supptable

\caption{\rev{Signal-derived probing: participant-mean $R^2$ [95\% CI]. N170 ($n=40$) targets PO8 amplitude; OpenBMI ($n=54$) targets log band power. Random averages three initialization seeds within participant.}}
\label{tab:physiology}
\begin{tabular*}{\linewidth}{@{\extracolsep{\fill}}lcccc@{}}
\toprule
& \multicolumn{2}{c}{N170} & \multicolumn{2}{c}{OpenBMI} \\
\cmidrule(lr){2-3}\cmidrule(l){4-5}
Input & Pretrained & Random & Pretrained & Random \\
\midrule
\multicolumn{5}{@{}l}{\textbf{A. Matched PCA200}} \\[3pt]
Raw EEG & \multicolumn{2}{c}{\suppcistack{0.864}{0.840}{0.884}} & \multicolumn{2}{c}{\suppcistack{-4.246}{-6.038}{-2.692}} \\
\addlinespace[2pt]
\addlinespace[3pt]
H01 & \suppcistack{0.628}{0.571}{0.678} & \suppcistack{0.474}{0.400}{0.536} & \suppcistack{0.402}{0.250}{0.535} & \suppcistack{-0.661}{-1.061}{-0.318} \\
\addlinespace[2pt]
H02 & \suppcistack{0.607}{0.536}{0.666} & \suppcistack{0.452}{0.376}{0.515} & \suppcistack{0.435}{0.327}{0.535} & \suppcistack{-0.721}{-1.142}{-0.357} \\
\addlinespace[2pt]
H03 & \suppcistack{0.598}{0.540}{0.650} & \suppcistack{0.430}{0.353}{0.494} & \suppcistack{0.366}{0.258}{0.468} & \suppcistack{-0.728}{-1.148}{-0.366} \\
\addlinespace[2pt]
\addlinespace[3pt]
H04 & \suppcistack{0.616}{0.560}{0.667} & \suppcistack{0.419}{0.338}{0.484} & \suppcistack{0.229}{0.089}{0.357} & \suppcistack{-0.746}{-1.163}{-0.382} \\
\addlinespace[2pt]
H05 & \suppcistack{0.587}{0.522}{0.645} & \suppcistack{0.397}{0.311}{0.465} & \suppcistack{0.245}{0.120}{0.362} & \suppcistack{-0.764}{-1.187}{-0.396} \\
\addlinespace[2pt]
H06 & \suppcistack{0.522}{0.436}{0.597} & \suppcistack{0.394}{0.313}{0.460} & \suppcistack{0.331}{0.224}{0.433} & \suppcistack{-0.770}{-1.189}{-0.403} \\
\addlinespace[2pt]
\addlinespace[3pt]
H07 & \suppcistack{0.453}{0.365}{0.530} & \suppcistack{0.374}{0.284}{0.446} & \suppcistack{0.288}{0.168}{0.400} & \suppcistack{-0.773}{-1.192}{-0.406} \\
\addlinespace[2pt]
H08 & \suppcistack{0.434}{0.336}{0.518} & \suppcistack{0.372}{0.281}{0.443} & \suppcistack{0.288}{0.177}{0.394} & \suppcistack{-0.780}{-1.200}{-0.414} \\
\addlinespace[2pt]
H09 & \suppcistack{0.306}{0.190}{0.408} & \suppcistack{0.371}{0.284}{0.441} & \suppcistack{0.254}{0.130}{0.370} & \suppcistack{-0.797}{-1.225}{-0.424} \\
\addlinespace[2pt]
\addlinespace[3pt]
H10 & \suppcistack{0.084}{-0.069}{0.220} & \suppcistack{0.375}{0.286}{0.446} & \suppcistack{0.287}{0.185}{0.386} & \suppcistack{-0.798}{-1.224}{-0.426} \\
\addlinespace[2pt]
H11 & \suppcistack{0.027}{-0.162}{0.187} & \suppcistack{0.376}{0.285}{0.447} & \suppcistack{0.293}{0.187}{0.396} & \suppcistack{-0.807}{-1.238}{-0.431} \\
\addlinespace[2pt]
H12 & \suppcistack{0.141}{-0.029}{0.283} & \suppcistack{0.366}{0.271}{0.440} & \suppcistack{0.239}{0.121}{0.354} & \suppcistack{-0.820}{-1.260}{-0.437} \\
\addlinespace[2pt]
\midrule
\multicolumn{5}{@{}l}{\textbf{B. Native controls: N170 5,600; OpenBMI 49,600 features}} \\[3pt]
Raw EEG & \multicolumn{2}{c}{\suppcistack{0.999}{0.998}{0.999}} & \multicolumn{2}{c}{\suppcistack{-9.049}{-11.091}{-7.213}} \\
\addlinespace[2pt]
\addlinespace[3pt]
H01 & \suppcistack{0.876}{0.856}{0.894} & \suppcistack{0.574}{0.513}{0.628} & \suppcistack{0.665}{0.610}{0.717} & \suppcistack{-0.448}{-0.739}{-0.195} \\
\addlinespace[2pt]
H12 & \suppcistack{0.181}{0.054}{0.297} & \suppcistack{0.462}{0.392}{0.526} & \suppcistack{0.486}{0.422}{0.550} & \suppcistack{-0.674}{-0.977}{-0.404} \\
\addlinespace[2pt]
\bottomrule
\end{tabular*}
\par\smallskip
\begin{minipage}{\linewidth}\footnotesize
Raw OpenBMI selects the largest Ridge penalty ($\alpha=100$) in every outer fold for both transforms. Negative $R^2$ values are retained. Intervals condition on the fitted probes and selected interfaces; they do not include full-pipeline retraining uncertainty.
\end{minipage}
\end{table}

%% file: tables/extra_adaptation.tex
\begin{table}[H]
\centering\color{RevisionBlue}\supptable

\caption{\rev{All-block LoRA on CBraMod/N170: native-dimensional diagnostic AUROC and paired differences (\%) [95\% participant-bootstrap CI], $n=40$. Aggregate scores average seeds within participant.}}
\label{tab:adaptation}
\supppanel{A. Freshly fitted diagnostic probes}
\begin{tabular*}{\linewidth}{@{\extracolsep{\fill}}lrrrr@{}}
\toprule
& \multicolumn{2}{c}{Frozen pretrained} & \multicolumn{2}{c}{All-block LoRA} \\
\cmidrule(lr){2-3}\cmidrule(l){4-5}
Diagnostic & AUROC & 95\% CI & AUROC & 95\% CI \\
\midrule
Final & 68.77 & [65.93, 71.67] & 79.57 & [76.21, 82.85] \\
First & 78.16 & [74.90, 81.37] & 77.84 & [74.43, 81.17] \\
Fixed selected & 77.45 & [74.25, 80.62] & 78.59 & [75.18, 81.91] \\
Reselected & --- & --- & 79.54 & [76.16, 82.83] \\
\bottomrule\end{tabular*}
\par\vspace{9pt}
\supppanel{B. Paired contrasts}
\begin{tabular*}{\linewidth}{@{\extracolsep{\fill}}lrr@{}}
\toprule
Contrast & Difference & 95\% CI \\
\midrule
Frozen: Fixed selected minus Final & 8.68 & [6.80, 10.43] \\
LoRA: First minus Final & -1.73 & [-2.45, -1.04] \\
LoRA: Fixed selected minus Final & -0.98 & [-1.73, -0.29] \\
LoRA: Reselected minus Final & -0.03 & [-0.18, 0.10] \\
LoRA: Reselected minus First & 1.70 & [1.00, 2.43] \\
LoRA Final minus Frozen Final & 10.80 & [8.67, 12.78] \\
LoRA Fixed minus Frozen Fixed & 1.14 & [0.29, 2.00] \\
\bottomrule\end{tabular*}
\par\vspace{9pt}
\supppanel{C. Individual optimization seeds}
\begin{tabular*}{\linewidth}{@{\extracolsep{\fill}}lrrrrrr@{}}
\toprule
Seed & Final & First & \shortstack{Fixed\\selected} & Reselected & \shortstack{Reselected\\minus Final} & 95\% CI \\
\midrule
1 & 79.71 & 77.80 & 78.61 & 79.62 & -0.09 & [-0.23, 0.04] \\
2 & 79.56 & 77.82 & 78.57 & 79.47 & -0.09 & [-0.27, 0.09] \\
3 & 79.45 & 77.90 & 78.60 & 79.52 & 0.07 & [-0.13, 0.28] \\
\bottomrule\end{tabular*}
\par\smallskip
\begin{minipage}{\linewidth}\footnotesize
The adaptation classifier, evaluated separately from these diagnostic probes, reaches 80.00 [76.64, 83.25]\% AUROC. Fixed selected uses H08/H02/H04/H02/H04 in outer folds 1--5; Reselected uses adapted inner validation. The pretrained Reselected entry is not a separate condition.
\end{minipage}
\end{table}

%% file: tables/extra_pretraining.tex
\begin{table}[H]
\centering\color{RevisionBlue}\supptable

\caption{\rev{Shared-backbone pretraining: absolute scores (\%) and fixed-readout depth gains (\%) [95\% CI], averaged over three paired seeds. The random encoder uses the same backbone.}}
\label{tab:pretraining}
\begin{tabular*}{\linewidth}{@{\extracolsep{\fill}}llrrr@{}}
\toprule
Task & Scheme & Final & Selected & Depth gain \\
\midrule
OpenBMI & MAE & 67.91 & 71.62 & \suppci{3.71}{2.71}{4.72} \\
 & JEPA & 67.06 & 71.66 & \suppci{4.60}{3.83}{5.35} \\
 & RANDOM & 70.67 & 70.67 & \suppci{0.00}{-0.24}{0.25} \\
\addlinespace[4pt]
N170 & MAE & 69.42 & 74.42 & \suppci{4.99}{3.62}{6.33} \\
 & JEPA & 63.82 & 74.47 & \suppci{10.65}{8.23}{12.91} \\
 & RANDOM & 70.59 & 70.47 & \suppci{-0.13}{-0.34}{0.09} \\
\addlinespace[4pt]
N400 & MAE & 73.47 & 74.70 & \suppci{1.23}{0.57}{1.86} \\
 & JEPA & 69.99 & 74.54 & \suppci{4.55}{3.29}{5.93} \\
 & RANDOM & 73.74 & 73.78 & \suppci{0.04}{-0.12}{0.19} \\
\bottomrule
\end{tabular*}
\end{table}

%% file: tables/extra_pretraining_gains.tex
\begin{table}[H]
\centering\color{RevisionBlue}\supptable

\caption{\rev{JEPA minus MAE differences in gains (\%) [95\% CI]. Larger gains do not imply higher absolute scores.}}
\label{tab:pretraining_gains}
\begin{tabular*}{\linewidth}{@{\extracolsep{\fill}}lrr@{}}
\toprule
Task & Difference in depth gain & Difference in residual gain \\
\midrule
OpenBMI & \suppci{0.89}{-0.06}{1.83} & \suppci{0.30}{-0.39}{0.99} \\
N170 & \suppci{5.66}{3.96}{7.19} & \suppci{9.65}{6.62}{12.23} \\
N400 & \suppci{3.32}{2.22}{4.48} & \suppci{3.20}{2.35}{4.10} \\
\bottomrule
\end{tabular*}
\end{table}